\pdfoutput=1

\documentclass[11pt]{article}

\usepackage{EMNLP2023}

\usepackage{times}
\usepackage{latexsym}
\usepackage{placeins}

\usepackage[T1]{fontenc}

\usepackage[utf8]{inputenc}

\usepackage{microtype}

\usepackage{inconsolata}

\usepackage{amsmath}
\usepackage{amsfonts}
\usepackage{hhline}
\usepackage{multirow}
\usepackage{makecell}

\usepackage[T1]{fontenc}    
\usepackage{url}            
\usepackage{booktabs, multirow, tabularx}       
\usepackage{amsfonts}       
\usepackage{nicefrac}       
\usepackage{microtype}      
\usepackage{graphicx}
\usepackage{subcaption}
\usepackage{natbib}
\usepackage{doi}
\usepackage{color}
\usepackage{amsmath}
\usepackage{mathtools, cuted}
\usepackage{etoolbox}
\usepackage{changepage}
\usepackage{pdflscape}
\usepackage{multicol}
\usepackage{balance}
\usepackage{wrapfig}
\usepackage{array}
\usepackage{soul}
\usepackage{bm}
\usepackage{enumitem}
\usepackage[symbol]{footmisc}
\usepackage[subtle]{savetrees}
\usepackage{longtable}

\newcolumntype{L}[1]{>{\raggedright\arraybackslash}p{#1}} 

\AfterEndEnvironment{strip}{\leavevmode}

\newcommand{\nce}{\mathrm{NCE}}

\newcommand{\JEmbeddingVThree}{\href{https://huggingface.co/jinaai/jina-embeddings-v3}{\texttt{jina-embeddings-v3}}}
\newcommand{\JEmbeddingVFour}{\href{https://huggingface.co/jinaai/jina-embeddings-v4}{\texttt{jina-embeddings-v4}}}
\newcommand{\jclipII}{\href{https://huggingface.co/jinaai/jina-clip-v2}{\texttt{jina-clip-v2}}}
\newcommand{\Qwen}{\texttt{Qwen2.5-VL-3B-Instruct}}
\newcommand{\JVDR}{Jina-VDR}

\newcommand{\rom}[1]{\uppercase\expandafter{\romannumeral #1\relax}}

\def\@fnsymbol#1{\ensuremath{\ifcase#1\or *\or \dagger\or \ddagger\or
   \mathsection\or \mathparagraph\or \|\or **\or \dagger\dagger
   \or \ddagger\ddagger \else\@ctrerr\fi}}
\newcommand{\ssymbol}[1]{^{\@fnsymbol{#1}}}

\newcommand{\softmax}{\mathrm{softmax}}
\definecolor{darkgreen}{rgb}{0.0, 0.5, 0.1}

\newcommand{\NA}{}

\title{\JEmbeddingVFour: Universal Embeddings for \\Multimodal Multilingual Retrieval}

\author{xxx$^*$, xxx$^*$, \\
Jina AI GmbH, Prinzessinnenstraße 19--20, 10969 Berlin, Germany \\
\texttt{research@jina.ai}
}

\author{Michael G\"unther$^*$, Saba Sturua$^*$, Mohammad Kalim Akram$^*$,   \\ \textbf{Isabelle Mohr$^*$}, \textbf{Andrei Ungureanu$^*$}, \textbf{Bo Wang$^*$}, \textbf{Sedigheh Eslami}, \textbf{Scott Martens}, \\ \textbf{Maximilian Werk}, \textbf{Nan Wang} and \textbf{Han Xiao}\\
	Jina AI GmbH, Prinzessinnenstraße 19, 10969, Berlin, Germany \\
	\texttt{research@jina.ai}}
\date{2024/02/26}

\hypersetup{
pdftitle={Jina Embeddings V4: Universal Embeddings for Multimodal Multilingual Retrieval},
pdfauthor={xxx}
pdfkeywords={Embeddings, Multilingual Models, Token length, Semantic Textual Similarity, Information Retrieval, Text Retrieval},
}

\begin{document}
\maketitle

\def\thefootnote{*}\footnotetext{Equal contribution.}\def\thefootnote{\arabic{footnote}}

\begin{abstract}
We introduce \JEmbeddingVFour{}, a 3.8 billion parameter multimodal embedding model that unifies text and image representations through a novel architecture supporting both single-vector and multi-vector embeddings in the late interaction style.
The model incorporates task-specific Low-Rank Adaptation (LoRA) adapters to optimize performance across diverse retrieval scenarios, including query-document retrieval, semantic text similarity, and code search.
Comprehensive evaluations demonstrate that \JEmbeddingVFour{} achieves state-of-the-art performance on both single-modal and cross-modal retrieval tasks, with particular strength in processing {visually rich} content such as tables, charts, diagrams, and mixed-media formats.
To facilitate evaluation of this capability, we also introduce \JVDR{}, a novel benchmark specifically designed for visually rich image retrieval.
\end{abstract}

\section{Introduction}
\label{sec:introduction}

We present \JEmbeddingVFour{}, a multimodal embedding model capable of processing text and image data to produce semantic embedding vectors of varying lengths, optimized for a broad array of applications. It incorporates optimized LoRA adapters~\cite{DBLP:conf/iclr/HuSWALWWC22} for information retrieval and semantic text similarity. An adapter is also provided for programming language embeddings, technical question-answering, and natural language code retrieval. It also brings new functionality to processing {visually rich} images (also called {visual documents}), i.e., materials mixing texts and images, containing tables, charts, diagrams, and other kinds of common mixed media~\cite{ding2024}. We have also developed \JVDR{}, a new multilingual, multi-domain benchmark suite for a broad range of visual retrieval tasks, to evaluate the capabilities of \JEmbeddingVFour{}.

We discuss the challenges of developing a multimodal, multi-functional, state-of-the-art embedding model capable of handling texts in a variety of languages, including computer coding languages, images, and “visually rich” data. The resulting model, \JEmbeddingVFour{}, projects inputs from all modalities into a unified semantic space, minimizing or eliminating the “modality gap” that has troubled similar projects~\cite{liang2022mind}. In addition, we introduce {\JVDR{}}, an advanced benchmark for images like screenshots and scans of visually complex documents.

The major contributions of this work are as follows:

\begin{itemize}
\item We introduce a unified multi-task learning paradigm that jointly optimizes embedding models to represent texts and images as single- and multi-vector embeddings.
\item Building on work done for \JEmbeddingVThree{}, we train LoRA extensions to enhance support for specific domains and task types, achieving results comparable to specialized models.
\item We have made particularly strong progress in handling visually rich images, especially for tasks outside of the existing ViDoRe benchmark~\cite{faysse2024colpali}, which is limited to question-answering. \JEmbeddingVFour{} outperforms other multimodal models by a significant margin on this type of material and supports a much more diverse set of use scenarios.
\item We construct a multilingual, multi-domain benchmark for screenshot retrieval. In contrast to other retrieval benchmarks (i.e., \cite{faysse2024colpali,xiao2025mieb}) that focus on question answering and OCR-related tasks, we expand the scope of visual document benchmarking to multilingual retrieval, more query types, and a much more diverse array of materials, like maps, diagrams, advertisements, and other mixed media.
\end{itemize}

\section{Background}

The underlying principles behind embedding models are indifferent to data modality. An embedding model transforms digitally encoded objects into vectors in a high-dimensional embedding space such that some of the semantic features of the objects, depending on the model’s training regimen, correspond to subspaces in that embedding space. Objects with more such features in common will have corresponding vectors that are closer to each other by some metric (typically cosine similarity) than objects with fewer common features.

Individual models, however, only support the modalities for which they are designed and trained. Embedding models initially developed primarily to support natural language texts, like \JEmbeddingVThree{}~\cite{sturua2024jina}, but there are many image embedding models, and more recently, audio and video models. The semantic embedding paradigm can also encompass models that support more than one modality, like bimodal image-text models, including OpenAI’s CLIP~\cite{radford2021learning} and subsequent developments including \texttt{jina-clip}~\cite{koukounas2024jina}. The principal purpose of multimodal embedding models is to project objects from multiple modalities into the same semantic embedding space, so that, for example, a picture of a cat and a text discussing or describing a cat will correspond to relatively close embedding vectors.

Embedding models can also specialize in specific types of input within a single modality. There are text embedding models designed for programming code~\cite{liu2024codexembed}, legal texts~\cite{voyage2024law}, and other special domains. There is also recent work in specialized image embedding models designed to support “visually rich” data, such as screenshots, charts, and printed pages that combine text and imagery and have internal visual structure~\cite{faysse2024colpali, ma2024unifying}.

There are other dimensions of embedding model specialization as well. Models can be optimized for specific tasks, such as information retrieval, clustering, and classification~\cite{sturua2024jina}. They can also vary based on the nature of the embeddings they produce. Most are single-/dense vector models, generating one embedding vector for whatever input they are given. There are also multi-vector/late interaction models, such as ColBERT~\cite{khattab2020colbert} and ColPali~\cite{faysse2024colpali}. Late interaction is generally a more precise measure of semantic similarity for retrieval, but has significantly greater storage and computing costs.

Instead of specializing, \JEmbeddingVFour{} builds on a single base model to provide competitive performance as a text, image, and cross-modal embedding model with strong performance handling visually rich documents. It supports both single-vector and multi-vector and is optimized to provide embeddings of varying lengths. Furthermore, the model includes LoRA extensions that optimize it for specific application classes: information retrieval, multimodal semantic similarity, and computer code retrieval.

This single-model approach entails significant savings in practical use cases when compared to deploying multiple AI models for different tasks and modalities.

\section{Related Work}
\label{sec:related_work}

Transformer-based neural network architectures that generate semantic embeddings are well-established~\cite{reimers2019sentence}, and there is a sizable literature on training techniques for them. Multi-stage contrastive training~\cite{wang2022text}, and techniques for supporting longer texts~\cite{gunther2023jina2} are particularly relevant to this work. 

Compact embedding vectors bring valuable performance benefits to AI applications, and this motivates work in Matryoshka Representational Learning (MRL)~\cite{kusupati2022matryoshka} as a way to train models for truncatable embedding vectors.

Contrastive text-image training has led to ground-breaking results in zero-shot image classification and cross-modal retrieval in conjunction with dual encoder architectures like CLIP~\cite{radford2021learning}. However, recent work shows better performance from {vision-language models} (VLMs) like \Qwen{}~\cite{bai2025qw}. \citet{jiang2024e5} show that VLMs suffer less from a modality gap than dual encoder architectures.

In contrast to \cite{zhang2024gme}, \JEmbeddingVFour{} is trained on multilingual data, supports single as well as multi-vector retrieval, and does not require task-specific instructions.
Other VLM models are trained exclusively on data for text-to-image~\cite{faysse2024colpali, ma2024unifying} or text-to-text retrieval~\cite{jiang2024e5}.

Similarity scoring in late interaction models does not use simple cosine similarity.~\cite{khattab2020colbert} Instead, similarity is calculated asymmetrically over two sequences of token embeddings — a query and a document — by summing up the maximum cosine similarity values of each query token embedding to any of the token embeddings from the document. Thus, for query embedding $q$ and document embedding $p$, their late interaction similarity score $s_\mathrm{late}(q,p)$ is determined by:

\begin{flalign}\label{eq:late-interaction-unnormalized}
s_\mathrm{late}(q,p) = \sum_{i=1}^n\max_{j\in\{1,\dots,m\}} \bm{q}_i \cdot \bm{p_j}^T
\end{flalign}

\citet{faysse2024colpali} train a late-interaction embedding model to search document screenshots using text queries, performing significantly better than traditional approaches involving OCR and CLIP-style models trained on image captions. To show this, they introduce the ViDoRe (Vision Document Retrieval) benchmark. However, this benchmark is limited to question-answering tasks in English and French involving only charts, tables, and pages from PDF documents. \citet{xiao2025mieb} extend this benchmark to create {MIEB} (Massive Image Embedding Benchmark) by adding semantic textual similarity (STS) tasks for visually rich documents like screenshots.

\section{Model Architecture}
\label{sec:model-architecture}
\begin{figure*}[htbp]
\centering
\includegraphics[width=\linewidth]{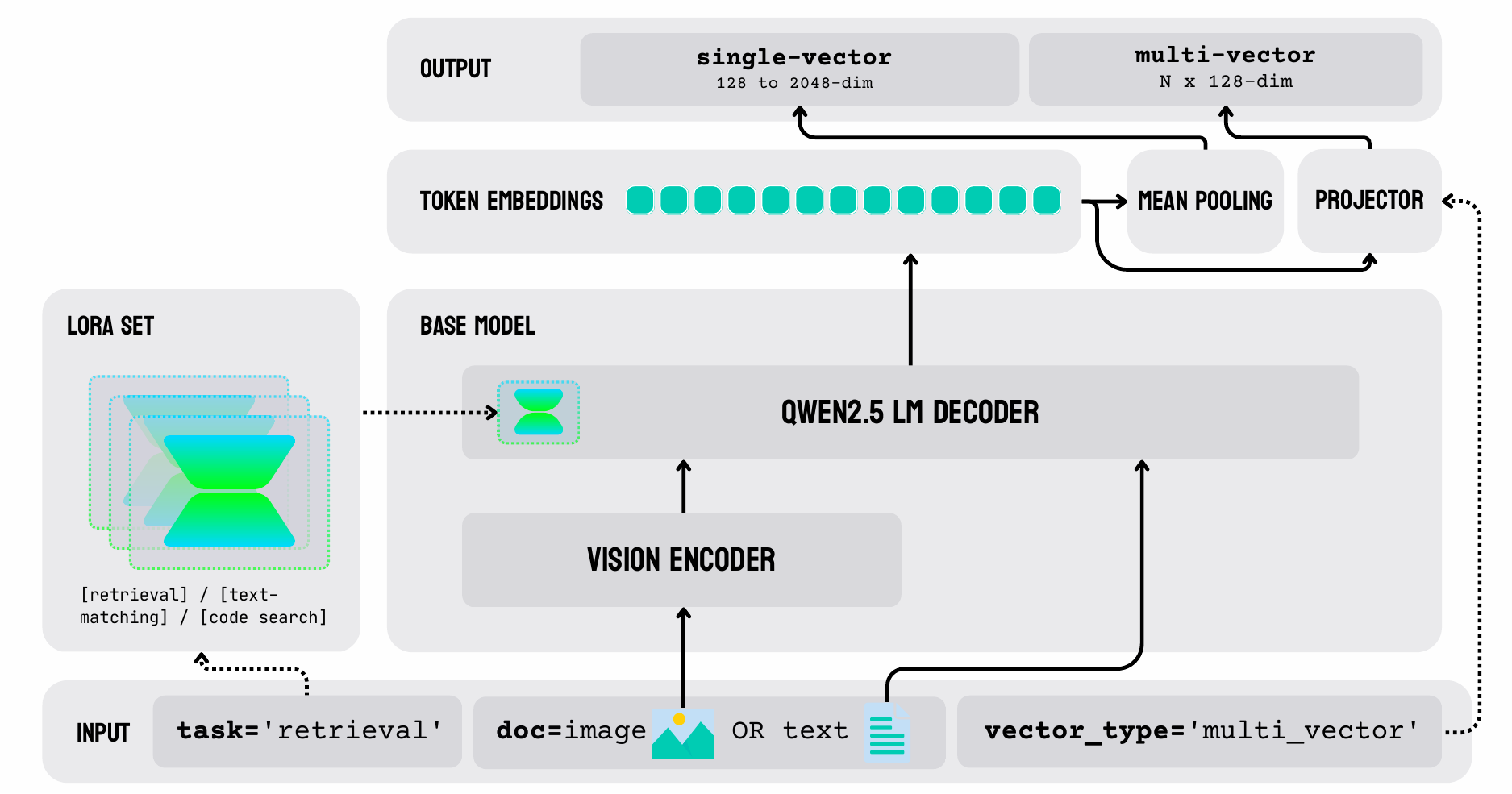}
\caption{Architecture of \JEmbeddingVFour{}. 
The model employs a unified LM built on the Qwen2.5-VL-3B-Instruct backbone (3.8B parameters). Text and image inputs are processed through a shared pathway: images are first converted to token sequences via a vision encoder, then both modalities are jointly processed by the language model decoder with contextual attention layers. Three task-specific LoRA adapters (60M parameters each) provide specialized optimization for retrieval, text-matching, and code search tasks without modifying the frozen backbone weights. The architecture supports dual output modes: (1) single-vector embeddings (2048 dimensions, truncatable to 128) generated via mean pooling for efficient similarity search, and (2) multi-vector embeddings (128 dimensions per token) via projection layers for the late interaction style retrieval.
}
\label{img:model-architecture}
\end{figure*}
The architecture of \JEmbeddingVFour{}, schematized in Figure~\ref{img:model-architecture}, 
employs a unified multimodal language model built on the \Qwen{}\footnote{\url{https://huggingface.co/Qwen/Qwen2.5-VL-3B-Instruct}} backbone~\cite{bai2025qw}. Text and image inputs are processed through a shared pathway: Images are first converted to token sequences via a vision encoder, then both modalities are jointly processed by the language model decoder with contextual attention layers. This unified design eliminates the modality gap present in dual-encoder architectures while maintaining competitive performance across text, image, and cross-modal tasks.

As shown in Figure~\ref{img:model-architecture}, this architecture supports dual output modes, as outlined in Section~\ref{sec:dual}. Furthermore, three task-specific LoRA adapters, each with 60M parameters, provide specialized task optimization without modifying the frozen backbone weights. These are described in Section~\ref{sec:lora}. 

The core specifications of \JEmbeddingVFour{} are summarized in Table~\ref{tab:specs}.

\begin{table}[t!]
\centering
\small{
\begin{tabular}{lp{4cm}}
\toprule
\texttt{Model Parameters} & 3.8 billion ($3.8 \times 10^9$) plus 60M per LoRA\\
\texttt{Text Input Size} & Up to 32,768 tokens \\ 
\texttt{Image Input}  &  All images resized to 20 megapixels \\
\makecell[l]{\texttt{Single-vector}\\ \texttt{Embedding Size}} & 2048 dimensions, truncatable down to 128 \\
\makecell[l]{\texttt{Multi-vector}\\ \texttt{Embedding Size}} & 128 dimensions per token \\
\bottomrule
\end{tabular}
}
\caption{Basic specifications of \JEmbeddingVFour{}}
\label{tab:specs}
\end{table}

\subsection{True Multimodal Processing}

The \Qwen{} paradigm differs from CLIP-style dual-encoder models in offering a single processing path that’s truly multimodal. 

For text input, \JEmbeddingVFour{} and \Qwen{} initially behave like other transformer-based embedding models: The text is tokenized, each token is replaced with a vector representation from a lookup table, and then these vectors are stacked and presented to a large language model (LLM). 

In CLIP-style models, images are processed by a separate embedding model, typically a transformer-based model that divides them into patches and then processes them much like a text model. The text model and image model are aligned during training to produce similar embeddings for similar semantic content in the different media.

The \Qwen{} paradigm used in \JEmbeddingVFour{} also includes a discrete image model but uses it in a different way. It produces a multi-vector result, comparable to late interaction models, and then passes this output to the LLM, as if it were a sequence of vectorized text tokens. The image embedding model acts as a preprocessor for the LLM, converting the image into what amounts to a sequence of vectorized “image tokens.”

This approach, which is the core of \JEmbeddingVFour{}, beyond having performance advantages, makes it possible to pass a text prompt into the LLM along with an image. The VLM is truly multimodal, since it is one model supporting multiple data types in a single input field.

\subsection{Dual Mode Output}
\label{sec:dual}

In contrast to \Qwen{} and other embedding models in general, users can choose between two output options: Traditional single (dense) vector embeddings and ColBERT-style multi-vector embeddings for late interaction strategies.

Single-vector embeddings are 2048 dimensions, but can be truncated to as little as 128 with minimal loss of precision. \JEmbeddingVFour{} has been trained with {Matryoshka Representation Learning}~\cite{kusupati2022matryoshka}, so that the scalar values of single-vector embeddings are roughly ordered by semantic significance. Eliminating the least significant dimensions reduces precision very little.

Multi-vector embeddings are the unpooled result of processing tokens through a transformer model. They correspond to tokens as the model analyses them, given their context. The length of the output vector is proportionate to the number of input tokens (including “image tokens”), with each token corresponding to a 128-dimensional output vector. This output is directly comparable to the unpooled embeddings produced by ColBERT~\cite{khattab2020colbert} and ColPali~\cite{faysse2024colpali} and is intended for use in late interaction comparison strategies.

For single-vector embeddings, mean pooling is applied to the final layer of the base model to produce the output.
The model incorporates an additional layer to project the output of the base model into multi-vector outputs.

\subsection{Task Specialization with LoRA}
\label{sec:lora}

Following the methods used for \JEmbeddingVThree{}~\cite{sturua2024jina}, we have implemented three task-specific LoRA adapters for different information retrieval use cases:

\begin{itemize}
    \item Asymmetric Query-Document Retrieval
    \item Semantic Similarity and Symmetric Retrieval
    \item Code (i.e., computer programming language) Retrieval
\end{itemize}

Asymmetric retrieval means encoding queries and documents differently in order to improve retrieval for queries that are not structured like documents, i.e., short queries, questions, etc. This is in contrast to symmetric retrieval, which assumes a symmetry between query and document, and is used to find comparable content.

Each LoRA adapter set has only 60M parameters, so maintaining all three adds less than 2\% to the memory footprint of \JEmbeddingVFour{}. Users can select among them at inference time, and all three support image and text encoding.

See Section~\ref{sec:evaluation} for performance information about these adapters.

\section{Training Method}
\label{sec:training}
Before training, model weights are initialized to match \texttt{Qwen/Qwen2.5-VL-3B-Instruct}. The multi-vector projection layer and LoRA adapters are randomly initialized. The weights of the backbone model are not modified during the training process. The LoRA adapters modify the effect of the backbone model layers and the projection layer. Only the adapters are trained.

Training proceeds in two phases:

\begin{enumerate}
    \item A single LoRA adapter is trained using contrasting text pairs and text-image pairs. We use the contrastive InfoNCE~\citep{DBLP:journals/corr/abs-1807-03748} loss function to co-train for both single-vector and multi-vector similarity, as detailed in the section below. No task-specific training is performed at this stage.
    \item The resulting LoRA adapter is duplicated to create the three task-specific adapters, which are then trained individually with task-specific text triplets and text-image triplets.
\end{enumerate}

In both phases of training, we apply {Matryoshka loss}~\cite{kusupati2022matryoshka} to the base loss so that single-vector embeddings from \JEmbeddingVFour{} are truncatable.

\subsection{Pair Training}
\label{sec:embedding-training}

Initially, training is performed with a contrastive objective. Pairs of inputs are classed as related or unrelated, and the model learns to embed related items closely together and unrelated items further apart.

In each training step, we sample two different batches of training data:

\begin{itemize}
    \item A batch $\mathcal{B}_{text}$ of text pairs.
    \item A batch $\mathcal{B}_{multi}$ of multimodal pairs containing a text and a related image.
\end{itemize}

We generate normalized single-vector and multi-vector embeddings for all texts and images in the selected pairs. We then construct a matrix of similarity values $\textbf{S}_\mathrm{dense}(\mathcal{B})$ by calculating the cosine similarity of all combinations of single-vector embeddings $\bm{q}_i$ and $\bm{p}_j$ in $\mathcal{B}$. We construct an analogous matrix $\textbf{S}_\mathrm{late}$ for each $\mathcal{B}$ for the multi-vector embeddings using a slightly modified version of Equation~\eqref{eq:late-interaction-unnormalized} to calculate their similarity. Our choice of loss function requires a normalized score, so we divide the late interaction score by the number of tokens in the query:

\begin{equation}\label{eq:late-interaction}
s'_\mathrm{late}(q_i,p_j) = \frac{s_\mathrm{late}(q_i,p_j)}{t}
\end{equation}

where $t$ is the number of tokens in $q_i$ and $q_i,p_j \in \mathcal{B}$

This modification is only for training. For retrieval applications, normalization is not necessary since the query is invariant.

Then, we apply the contrastive InfoNCE loss function $\mathcal{L}_{\mathrm{NCE}}$~\citep{DBLP:journals/corr/abs-1807-03748} on each of the four resulting matrices of similarity scores $s_{i,j} \in \textbf{S}$

\begin{equation}
    \softmax(\textbf{S},\tau,i,j) := \ln \frac{e^{s_{i,j} / \tau}}{\sum\limits_{k = 0}^n e^{s_{i,k} / \tau}}
\end{equation}

\begin{equation}
    \label{eq:info-nce}
    \mathcal{L}_{\nce}(\textbf{S}(\mathcal{B}),\tau) := -\sum_{i,j=0}^{n} \softmax(\textbf{S}(\mathcal{B}),\tau,i,i)
\end{equation}

where $\tau$ is the {temperature} parameter, $n$ is the batch size, which increases the weight of small differences in similarity scores in calculating the loss.

Following \citet{hinton2015}, we compensate for differences in error distributions between the single-vector and multi-vector by adding the weighted Kullback–Leibler divergence ($D_{KL}$) of the two sets of softmax-normalized similarity scores. This enables us to train for the single-vector and multi-vector outputs simultaneously, even though the multi-vector/late interaction scores have much less error.

\begin{flalign}
    \mathcal{L}_D(B,\tau)& :=D_\mathrm{KL}(\textbf{S}'_\mathrm{dense}(\mathcal{B}) \,\|\, \textbf{S}'_\mathrm{late}(\mathcal{B})) \nonumber\\
    \text{where}\;\;\;\; & 
    \textbf{S}'_{i,j} = \softmax(\textbf{S},\tau,i,j)
\end{flalign}

The resulting joint loss function, which we use in training, is defined as:

\begin{flalign}
    \mathcal{L}_\mathrm{joint}&(\mathcal{B}_\mathrm{txt}, \mathcal{B}_\mathrm{multi},\tau) := \nonumber\\
    &w_1\mathcal{L}_{\nce}( \textbf{S}_\mathrm{dense}(\mathcal{B}_\mathrm{txt}), \tau) \nonumber\\ 
    + &w_2\mathcal{L}_{\nce}( \textbf{S}_\mathrm{late}(\mathcal{B}_\mathrm{txt}), \tau) + w_3\mathcal{L}_D(\mathcal{B}_\mathrm{txt}) \nonumber\\
    + &w_4\mathcal{L}_{\nce}( \textbf{S}_\mathrm{dense}(\mathcal{B}_\mathrm{multi}), \tau) \nonumber\\
    + &w_5\mathcal{L}_{\nce}( \textbf{S}_\mathrm{late}(\mathcal{B}_\mathrm{multi}), \tau) + w_6\mathcal{L}_D(\mathcal{B}_\mathrm{multi}) 
\end{flalign}

The weights $w_1, \ldots, w_6$ and temperature $\tau$ are training hyperparameters.

\subsubsection{Pair Training Data}
The training data consists of text-text and text-image pairs from more than 300 sources. Text-text pairs are selected and filtered as described in \citet{sturua2024jina}. Text-image pairs have been curated from a variety of sources following a more eclectic strategy than previous work on training text-image embedding models. In contrast to relying on image-caption pairs or pairs of queries and images derived from documents, we have also created images from other document types. Our training data includes website screenshots, rendered Markdown files, charts, tables, and other kinds of materials "found in the wild." The queries consist primarily of questions, keywords and key phrases, long descriptions, and statements of fact.

\subsection{Task-Specific Training}
\label{sec:supervised_fine_tuning}
We instantiate three copies of the pair-trained LoRA adapter and give each specific training for its intended task. Training data and loss functions differ for the three tasks.

\begin{table}[t!]
\centering
\small{
\begin{tabular}{lp{4cm}}
\toprule
Task Name      & Description \\
\midrule
\texttt{retrieval}        & Asymmetric embedding of queries and documents for retrieval \\ 
\texttt{text-matching}            &  Semantic text similarity and symmetric retrieval \\
\texttt{code}           & Retrieving code snippets \\ 
\bottomrule
\end{tabular}
}
\caption{Supported tasks of \JEmbeddingVFour{}, each corresponding to a LoRA adapter and trained independently}
\label{tab:task-value}
\end{table}

\subsubsection{Asymmetric Retrieval Adapter}
\label{sec:retrieval-adapter}

Asymmetric retrieval assigns substantially and qualitatively different embeddings to documents and queries, even if they happen to have the very same text. Having distinct encoding mechanisms for the two often significantly benefits embeddings-based retrieval performance. \citet{sturua2024jina} shows that this can be achieved either by training two separate adapters or by employing two distinct prefixes as proposed in \citet{wang2022text}, so that embedding models can readily distinguish them when they generate embeddings.

We have used the prefix method for \JEmbeddingVFour{}. Previous work shows little benefit from combining both methods.

Our training data contains {hard negatives}, i.e., triplets of a query, a document that matches the query, and a document that is closely semantically related but not a correct match~\cite{wang2022text, li2023general, gunther2023jina2}. For every pair $(q_i,p_i) \in \mathcal{B}$ in a batch, $p_i$ is intended to be a good match for $q_i$, and we presume that for all $(q_j,p_j) \in \mathcal{B}$ where $j \neq i$, $p_j$ is a bad match for $q_i$.

 We incorporate those additional negatives into the training process via an extended version of the $\mathcal{L}_{\nce}$ loss described in~\citet{gunther2023jina2}, denoted as $\mathcal{L}_{\mathrm{NCE+}}$, in our joint loss function $\mathcal{L}_\mathrm{joint}$:

\begin{flalign}
&\mathcal{L}_{\mathrm{NCE+}}(\textbf{S}(\mathcal{B}), \tau) := \nonumber\\
&\;\;\;\;\;\sum_{r\in \mathcal{B}}\Bigg[-\ln \frac{e^{s(q, p) / \tau}}{\sum\limits_{i = 1}^k \Big[ e^{s(q, p_i) / \tau}+ \sum\limits_{j = 1}^{m} e^{s(q, n_{j,i}) / \tau}\Big]}\Bigg]\nonumber \\
\label{eq:loss-multi-negatives}
\end{flalign}

with $r = (q,p, n_1, \ldots, n_{m})$, where $(q,p)$ is a pair in batch $\mathcal{B}$ and $n_1, \ldots, n_{m}$ and the other $p \in \mathcal{B}$.

Our dataset of text hard negatives is similar to the data used to train \JEmbeddingVThree{}~\cite{sturua2024jina}.
We rely on existing datasets to create multimodal hard negatives for training, including Wiki-SS~\cite{ma2024unifying} and VDR multilingual\footnote{\url{https://huggingface.co/datasets/llamaindex/vdr-multilingual-train}}, but also mined hard negatives from curated multimodal datasets.

\subsubsection{Text Matching Adapter}
\label{sec:text-matching}

Symmetric semantic similarity tasks require different training from asymmetric retrieval. We find that training data with ground truth similarity values works best for this kind of task. As discussed in~\citet{sturua2024jina}, we use the CoSENT\footnote{\url{https://github.com/bojone/CoSENT}} loss function
$\mathcal{L}_\mathrm{co}$ from ~\citet{li2024aoe}:

\begin{flalign}
\label{eq:cosentloss}
    &\mathcal{L}_{\mathrm{co}}(\textbf{S}(\mathcal{B}), \tau) := \ln \Big[  1 + \sum\limits_{\substack{(q_1,p_1), \\ (q_2,p_2) \\ \in \textbf{S}(\mathcal{B})}} \frac{e^{s(q_2,p_2)} - e^{s(q_1,p_1)}}{\tau} \Big] && \nonumber \\ 
\end{flalign}

 where $\zeta(q,p)$ is the ground truth semantic similarity of $q$ with $p$,  $\zeta(q_1, p_1) > \zeta(q_2, p_2)$, and $\tau$ is the temperature parameter.

The loss function operates on two pairs of text values, $(q_1, p_1)$ and $(q_2, p_2)$, with known ground truth similarity.

To train the model with this objective, we use data from semantic textual similarity (STS) training datasets such as STS12~\cite{agirre2012semeval} and SICK~\cite{marelli-etal-2014-sick}. The amount of data in this format is limited, so we enhance our ground truth training data with pairs that do not have known similarity scores. For these pairs, we proceed the same way as we did for pair training in Section~\ref{sec:embedding-training} and use the standard InfoNCE loss from Equation~\eqref{eq:info-nce}. The joint loss function is calculated as in Equation~\eqref{eq:loss-multi-negatives} except the CoSENT loss is used where pairs with known ground truth values exist.

\subsubsection{Code Adapter}
\label{sec:code-adapter}

Code embeddings in \JEmbeddingVFour{} are designed for natural language-to-code retrieval, code-to-code similarity search, and technical question answering. Code is a very specialized kind of text and requires distinct data sources. Because code embeddings do not involve image processing, the vision portion of \JEmbeddingVFour{} is not affected by training the code retrieval LoRA adapter.

The backbone LLM \Qwen{} was pre-trained on data including the StackExchangeQA\footnote{\url{https://github.com/laituan245/StackExchangeQA}} and the CodeSearchNet~\cite{husain2020codesearchnet} datasets, giving it some capacity to support code embeddings before further adaptation. Our LoRA training used the same triplet-based method described in Section~\ref{sec:retrieval-adapter}. Training triplets are derived from a variety of sources, including CodeSearchNet, CodeFeedback~\cite{zheng2024opencodeinterpreter}, APPS~\cite{hendrycks2021apps}, and the CornStack dataset~\cite{suresh2025cornstack}.

We maintained a consistent training configuration by using the same input prefix tokens (e.g., query, passage) and temperature hyperparameter (set to $0.02$) during the triplet-based training.

\section{\JVDR{}: Visually Rich Document Retrieval Benchmark}
\label{sec:benchmark}

To evaluate the performance of \JEmbeddingVFour{} across a broad range of visually rich document retrieval tasks, we have produced a new benchmark collection and released it to the public.\footnote{Benchmark available at \url{https://huggingface.co/collections/jinaai/jinavdr-visual-document-retrieval-684831c022c53b21c313b449}.} 

This new collection tests an embedding model's ability to integrate textual and visual understanding of documents that consist of in the form of rendered images of visual elements like charts, tables, and running text.
It extends the ViDoRe benchmark~\cite{faysse2024colpali} by adding a diverse collection of datasets spanning a broad range of domains (e.g. legal texts, historic documents, marketing materials), covering a variety of material types (e.g. charts, tables, manuals, printed text, maps) and query types (e.g. questions, facts, descriptions), as well as multiple languages.

The benchmark suite encompasses ViDoRe and adds 30 additional tests. These tests include re-purposed existing datasets, new manually-annotated datasets, and generated synthetic data 

For a comprehensive overview of the individual benchmarks, see Appendix~\ref{app:dataset-collection}.

\subsection{Re-purposed Datasets}

We have adapted a number of existing VQA and OCR datasets, modifying and restructuring them into appropriate query-document pairs.

For example, for {DonutVQA}\footnote{\url{https://huggingface.co/datasets/warshakhan/donut_vqa_ISynHMP}}, {TableVQA}~\cite{AgDeTQA}, {MPMQA}~\cite{zhang2023mpmqa}, {CharXiv}~\cite{wang2024charxiv}, and {PlotQA}~\cite{Methani_2020_PLOTQA}, we used structured templates and generative language models to formulate text queries to match their contents. 

{JDocQAJP}\footnote{\url{https://huggingface.co/datasets/jlli/JDocQA-nonbinary}} and {HungarianDocQA}\footnote{\url{https://huggingface.co/datasets/jlli/HungarianDocQA-OCR}} already contain documents and queries in forms that require minimal processing to adapt as benchmarks.

We also created datasets from available data that extend beyond conventional question formats. The {OurWorldInData} and {WikimediaMaps} datasets use encyclopedia article fragments and image descriptions as queries to match with charts and maps. The {GitHubREADMERetrieval} dataset contains rendered Markdown pages drawn from GitHub README files, paired with generated natural language descriptions in 17 languages. The {WikimediaCommonsDocuments} benchmark pairs multilingual document pages with paragraph-level references extracted from Wikipedia.

\subsection{Manually Annotated Datasets}

We have curated a number of human-annotated resources to better reflect real-world use cases. These include academic slides from Stanford lectures~\cite{mitchener2021banking}, educational figures in the {TQA} dataset~\cite{tqa}, and marketing and institutional documents such as the {Jina AI 2024 Yearbook}~\cite{jina2024research}, {Japanese Ramen}~\cite{niigata2024ramen}, and the {Shanghai Master Plan}~\cite{shanghai_masterplan_2018}. Documents in these datasets were paired with carefully written human queries without template-based phrasing, capturing genuine information-seeking intent. Some of these datasets target specific languages and regions to provide broader coverage.

We also incorporated pre-existing human-annotated datasets like {ChartQA}\footnote{\url{https://huggingface.co/datasets/HuggingFaceM4/ChartQA}} and its Arabic counterpart, {ArabicChartQA}~\cite{ghaboura2024camelbenchcomprehensivearabiclmm}, which focus on charts and infographics.

\subsection{Synthetic Data Generation}

We have been attentive, in constructing \JVDR{}, to the lack of diversity that often plagues information retrieval benchmarks. We cannot commission human-annotated datasets for everything and have had recourse to generative AI to fill in the gaps.

We obtained a number of datasets from primarily European sources containing scans of historical, legal, and journalistic documents in German, French, Spanish, Italian, and Dutch. We used Qwen2 to generate queries for these documents. We handled the {HindiGovernmentVQA} and {RussianBeverages} datasets in the same way, adding not only often underrepresented languages, but also public service documents and commercial catalogs to this benchmark set.

{TweetStockRetrieval}\footnote{\url{https://www.kaggle.com/datasets/thedevastator/tweet-sentiment-s-impact-on-stock-returns}} is a collection of chart-based financial data, which we have paired with multilingual template-based generated queries. {AirBnBRetrieval}\footnote{\url{https://www.kaggle.com/datasets/dgomonov/new-york-city-airbnb-open-data}} is a collection of rendered tables that we have paired with queries in 10 languages generated from a template.

In several cases, such as {TableVQA}~\cite{DeAFTdb}, we introduced bilingual examples (e.g., French/English) to better assess cross-lingual retrieval performance, with questions and answers synthesized using advanced multilingual LLMs such as Gemini 1.5 Pro and Claude 3.5 Sonnet.

\subsection{\JVDR{} in a Nutshell}

\JVDR{} extends the ViDoRe benchmark with:
\begin{itemize}
    \item 30 new tasks, using both real-world and synthetic data
    \item All datasets adapted for retrieval and designed to be compatible with ViDoRe
    \item LLM-based filtering to ensure all queries are relevant and reflective of real-world querying
    \item Non-question queries, such as GitHub descriptions matched to rendered markdown images, and map images from Wikimedia Commons with accompanying textual descriptions
    \item Multilingual coverage, with some datasets spanning up to 20 languages
\end{itemize}

\section{Evaluation}
\label{sec:evaluation}

We have evaluated \JEmbeddingVFour{} on a diverse set of benchmarks to reflect its multiple functions. Table~\ref{tab:eval-summary} provides an overview of benchmark averages for \JEmbeddingVFour{} and other embedding models.

\begin{table*}[ht]
{\small
\begin{center}
\caption{Average Retrieval Scores of Embedding Models on Various Benchmarks.}\label{tab:eval-summary}
\setlength{\tabcolsep}{3pt}
\begin{tabular}{lccccccccc}
\toprule
\textbf{Model} & \textbf{J-VDR} & \textbf{ViDoRe} & \textbf{CLIPB} & \textbf{MMTEB} & \textbf{MTEB-en} & \textbf{COIR} & \textbf{LEMB} & \textbf{STS-m} & \textbf{STS-en} \\
\midrule
jina-embeddings-v4 (dense) & 73.98 & 84.11 & 84.11 & 66.49 & 55.97 & 71.59  & 67.11 & 72.70 & 85.89 \\
jina-embeddings-v4 (late) & 80.55 &  90.17 & \NA & \NA & \NA & \NA & \NA & \NA  & \NA \\
\midrule
text-embedding-3-large & -- & -- & -- & 59.27 & 57.98 & 62.36 & 52.42 & 70.17 & 81.44 \\
bge-m3 & -- & -- & -- & 55.36 & \NA & \NA & 58.73 & \NA & \NA \\
multilingual-e5-large-instruct & -- & -- & -- & 57.12 & 53.47 & \NA & 41.76 & \NA & \NA \\
jina-embeddings-v3 & 47.82 & 26.02 & -- & 58.58 & 54.33 & 55.07 & 55.66 & 75.77 & 85.82 \\
voyage-3 & -- & -- & -- & 66.13 & 53.46 & 67.23 & 74.06 & 68.33 & 78.59 \\
gemini-embedding-001 & -- & -- & -- & 67.71 & 64.35 & 73.11 & \NA & 78.35 & 85.29 \\
jina-embedings-v2-code & -- & -- & -- & \NA & \NA & 52.24 & \NA & \NA & \NA \\
voyage-code & -- & -- & -- & \NA & \NA & 77.33 & \NA & \NA & \NA \\
\midrule
nllb-clip-large-siglip & \NA & \NA & 83.19 & \NA & \NA & \NA  & \NA & \NA & \NA \\
jina-clip-v2 & 40.52 & 53.61 & 81.12 & \NA & \NA & \NA & \NA & \NA & \NA \\
\midrule
colpali-v1.2 (late) & 63.80 & 83.90 & \NA & \NA & \NA & \NA & \NA & \NA & \NA \\
dse-qwen2-2b-mrl-v1 (dense) & 67.25 & 85.80 & \NA & \NA & \NA & \NA & \NA & \NA & \NA \\
voyage-multimodal-v3 (dense) & \NA & 84.24 & \NA & \NA & \NA & \NA & \NA & \NA & \NA \\
\bottomrule
\end{tabular}
\end{center}
\vspace{2mm}

\textbf{Task Acronyms:} J-VDR = Jina VDR, VidoRE = ViDoRe, CLIPB = CLIP Benchmark, MMTEB = MTEB(Multilingual, v2) Retrieval Tasks, MTEB-EN = MTEB(eng, v2) Retrieval Tasks, COIR = CoIR Code Retrieval, LEMB = LongEmbed, STS-m = MTEB(Multilingual, v2) Semantic Textual Similarity Tasks, STS-en = MTEB(eng, v2) Semantic Textual Similarity Tasks
\\ \textbf{Average Calculation:} For J-VDR and ViDoRE, we calculate the average for the multilingual tasks first and consider this as a single score before calculating the average across all tasks.
Scores are nDCG@5 for J-VDR, ViDoRe, and CLIPB, and nDCG@10 for MMTEB, MTEB-en, COIR, and LEMB, and Spearman coefficient for STS-m and STS-en.
\\ \textbf{Evaluation of Text Retrieval Models on J-VDR:} For evaluating text retrieval models on J-VDR, we used EasyOCR (\url{https://github.com/JaidedAI/EasyOCR}) and the provided extracted texts from the original ViDoRe datasets.
}
\end{table*}

\subsection{Multilingual Text Retrieval}

{MTEB} and {MMTEB}~\cite{enevoldsen2025mmtebmassivemultilingualtext} are the most widely used text retrieval benchmarks. For most tasks, we have used the asymmetric retrieval adapter, but for some symmetric retrieval tasks like {ArguAna}\footnote{\url{https://huggingface.co/datasets/mteb/arguana}}, we have used the text matching adapter instead. For evaluation, we prepend the query with the prefix ``Given a claim, find documents that refute the claim'' to reflect the task's focus on retrieving passages that contradict, rather than support, the input claim, similar to ~\citet{wang2023improving}. The results are tabulated in Appendix~\ref{app:mteb}.

For the MTEB benchmarks, which are all in English, 
see Table~\ref{tab:mteb_retrieval}, and for the multilingual MMTEB, see Table~\ref{tab:mmteb_retrieval}.
The performance of this new model is generally better than our previous model \JEmbeddingVThree{} and broadly comparable with the state-of-the-art. 

We have also evaluated the performance of our model on retrieval tasks that involve long text documents using the {LongEmbed} benchmark~\cite{zhu2024longembed}. The results are tabulated in Table~\ref{tab:longembed} of Appendix~\ref{app:mteb}. Long document performance for \JEmbeddingVFour{} significantly outpaces competing models except the \text{voyage-3} series and improves dramatically on \JEmbeddingVThree{}'s performance.

\subsection{Textual Semantic Similarity}

We evaluated \JEmbeddingVFour{} with text-based semantic similarity (STS) benchmarks. The results for MTEB STS and MMTEB STS benchmarks are tabulated in Tables~\ref{tab:mteb_sts} and~\ref{tab:mmteb_sts} of Appendix~\ref{app:mteb} respectively. Our results are competitive with the state-of-the-art and are best-in-class for English similarity tasks.

\subsection{Multimodal Retrieval}

To evaluate the model's performance on typical text-to-image search tasks, we used the common English and non-English tasks of the CLIP Benchmark\footnote{\url{https://github.com/LAION-AI/CLIP_benchmark}}. The results are tabulated in Tables~\ref{tab:clip} to \ref{tab:cm3600} of Appendix~\ref{app:clip}.
\JEmbeddingVFour{} has a higher average score than \jclipII{} and \texttt{nllb-siglip-large}, but the latter performs somewhat higher on the Crossmodal3600 benchmark~\cite{crossmodal2022} (see Table~\ref{tab:cm3600}) because it includes content from low-resource languages not supported in \JEmbeddingVFour{}'s \Qwen{} backbone.

We further tested \JEmbeddingVFour{} on the ViDoRe and \JVDR{} benchmarks, to evaluate its performance on visually rich documents. 
The results are compiled in Appendix~\ref{app:jvdr_results}.
ViDoRe scores are tabulated in Table~\ref{tab:vidore_retrieval}. 
Table~\ref{tab:jinavdr_overview} provides an overview of \JEmbeddingVFour{} compared to other models, with Tables~\ref{tab:wikicommons_results} to~\ref{tab:airbnb_results} providing details results for some individual \JVDR{} benchmarks.

This suggests that other models are primarily trained to perform well on document retrieval tasks that are similar to the ViDoRe tasks but underperform on other tasks, e.g., that do not involve queries that resemble questions.

\JEmbeddingVFour{} excels at this benchmark, providing the current state-of-the-art, in both single- and multi-vector mode. Multi-vector/late interaction matching is generally recognized as more precise than single-vector matching in other applications, and this remains true for \JVDR{}.

\subsection{Code Retrieval}

To assess performance on code retrieval, we evaluate the model on the MTEB-CoIR benchmark~\cite{li2024coir}, which consists of 10 tasks spanning text-to-code, code-to-text, code-to-code, and hybrid code retrieval types. The results are reported in Table~\ref{tab:code_retrieval} of Appendix~\ref{app:mteb}. \JEmbeddingVFour{} is competitive with the state-of-the-art in general-purpose embedding models, but the specialized \texttt{voyage-code} model has somewhat better benchmark performance.

\section{Analysis of the Embedding Space}

The large difference in architecture between \JEmbeddingVFour{} and CLIP-style models like OpenAI CLIP~\cite{radford2021learning} and \jclipII{} implies a large difference in the structure of the embedding spaces those models generate. We look here at a few of these issues.

\subsection{Modality Gap} 

Previous work has shed light on the so-called {modality gap} in multimodal models trained with contrastive learning~\cite{mind_the_gap, two_effects, alignclip}. Good semantic matches across modalities tend to lie considerably further apart in the embedding space than comparable or even worse matches of the same modality, i.e., texts in CLIP-style models are more similar to semantically unrelated texts than to semantically similar images.

\begin{figure}
  \begin{subfigure}[b]{0.5\textwidth}
  \includegraphics[width=\linewidth]{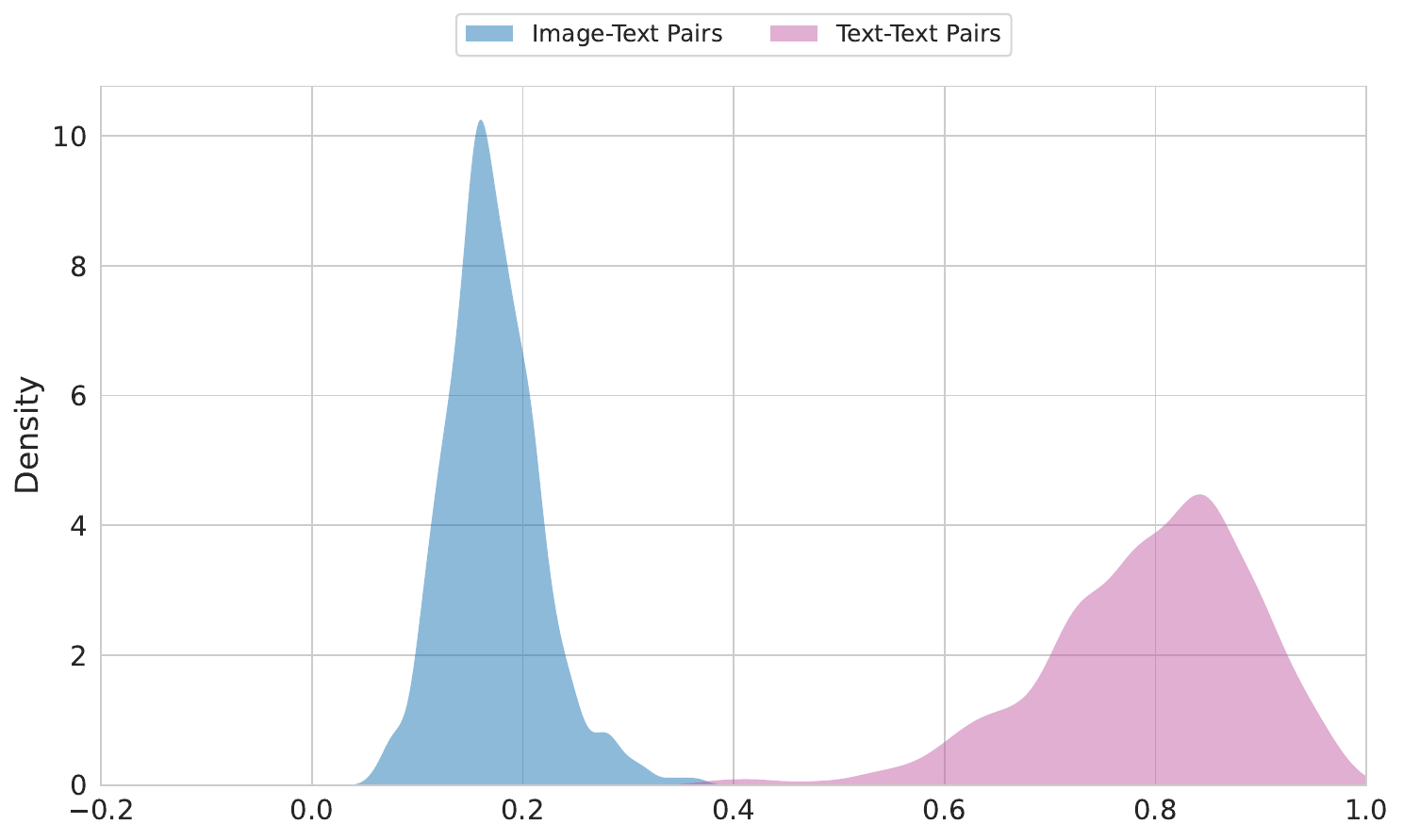}
  \end{subfigure}
  \begin{subfigure}[b]{0.5\textwidth}
  \includegraphics[width=\linewidth]{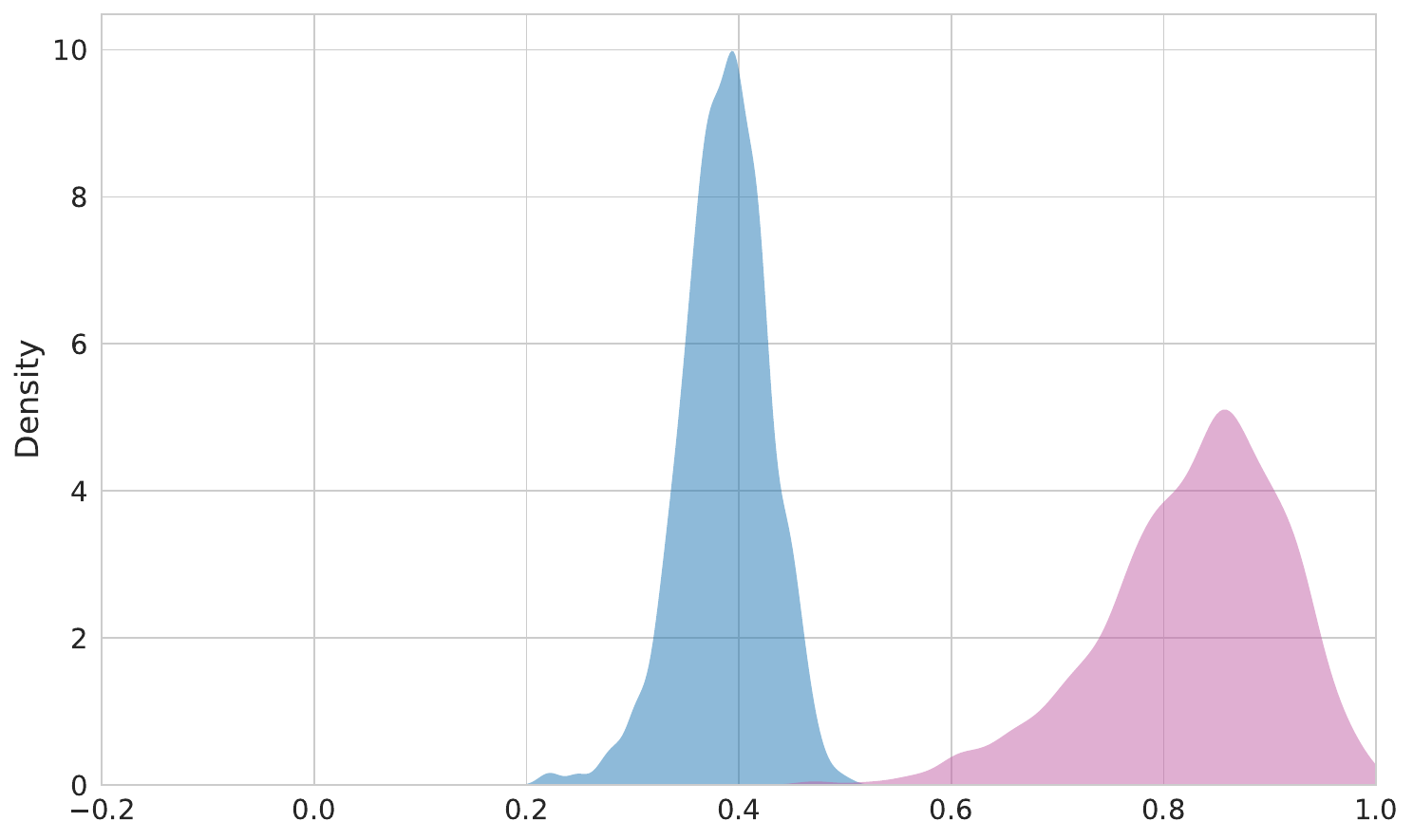}
  \end{subfigure} 
  \begin{subfigure}[b]{0.5\textwidth}
  \includegraphics[width=\linewidth]{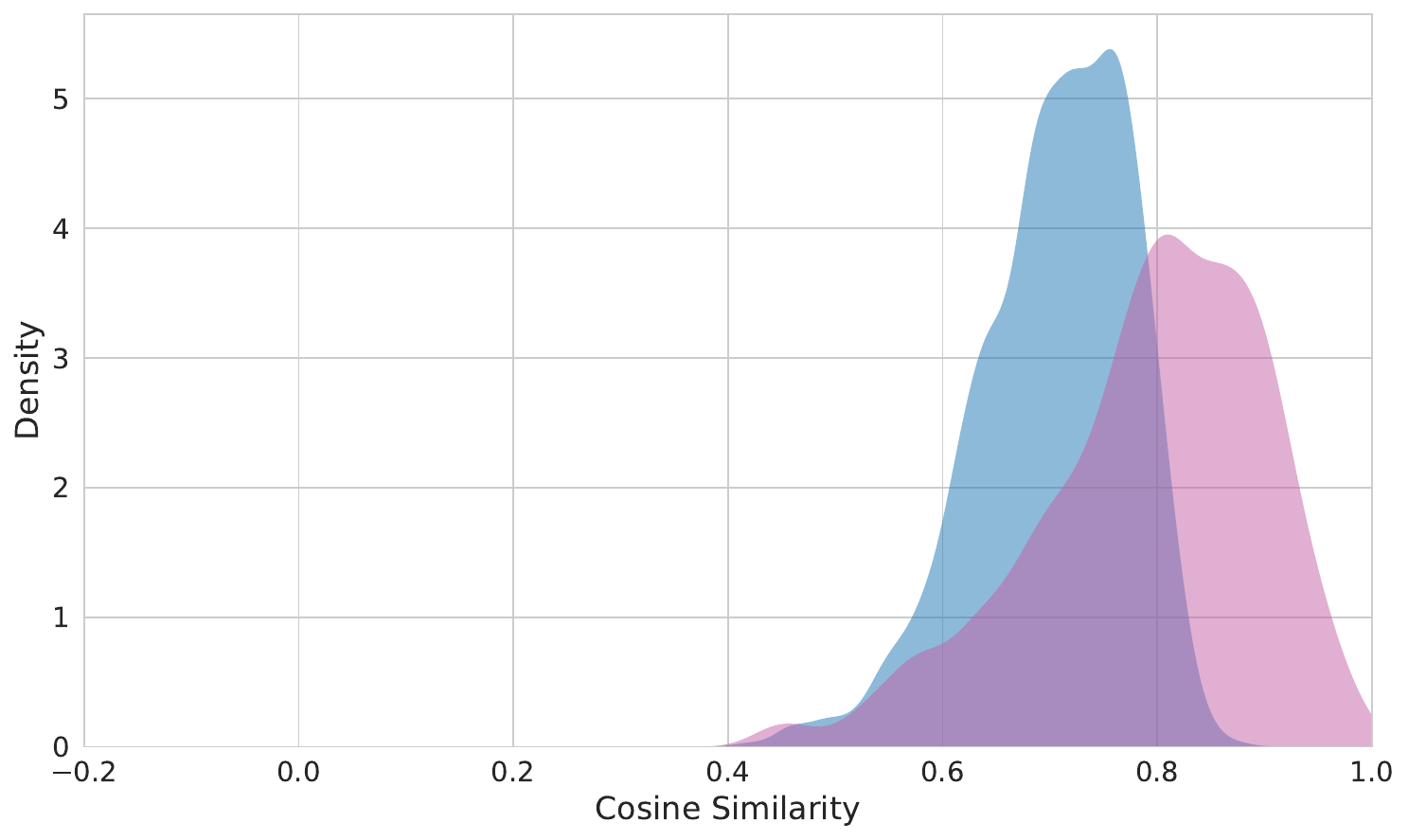}
  \end{subfigure} 
  \caption{Distribution of the cosine similarities of the paired image-text embeddings versus paired text-text embeddings from the Flickr8K\footnote{\url{https://www.kaggle.com/datasets/adityajn105/flickr8k}} dataset. \textbf{Top}: OpenAI CLIP, \textbf{Middle}: \jclipII, \textbf{Bottom}: \JEmbeddingVFour}
  \label{fig:modality-gap}
\end{figure}

We can see the modality gap directly by examining the distribution of pairwise cosine similarities of matching image-text pairs versus matching text-text pairs. In Figure~\ref{fig:modality-gap}, we see the distribution of similarity values for the two pair types in OpenAI CLIP, \jclipII{}, and \JEmbeddingVFour{}.

The gap is dramatically reduced with \JEmbeddingVFour{} because of its cross-model encoder, illustrated in Figure~\ref{img:model-architecture}. \citet{alignclip} has shown that sharing an encoder between modalities introduces an inductive bias towards using a shared region of the embedding space, while \citet{mind_the_gap} shows the opposite is true for CLIP-style architectures with separate encoders.

\subsection{Cross-Modal Alignment}

\citet{alignclip} has defined the {cross-modal alignment score} of a multimodal embedding model as the average of cosine similarities of matching pairs of image and text embeddings. Table~\ref{table:cross-modal-alignment} calculates this score for \JEmbeddingVFour{} and OpenAI CLIP with data sampled from the Flickr30K\footnote{\url{https://www.kaggle.com/datasets/adityajn105/flickr30k}}, MSCOCO~\cite{mscoco2014}, and CIFAR-100\footnote{\url{https://www.kaggle.com/datasets/fedesoriano/cifar100}} datasets.

These results confirm that \JEmbeddingVFour{} generates a far better aligned cross-modal embedding space than CLIP-style models.

It is worth noting that \JEmbeddingVFour{} shows much poorer alignment for CIFAR-100 data than MSCOCO and Flickr30K. This is because CIFAR-100 is a classification dataset and its labels are far less informative than the more descriptive texts in MSCOCO and Flickr30K.

\begin{table}[t]
\centering
\resizebox{\columnwidth}{!}{%
\begin{tabular}{lccc}
\toprule
\bfseries Model & \bfseries Flickr30K & \bfseries MSCOCO & 
\bfseries CIFAR-100 \\
\midrule
OpenAI-CLIP & 0.15 & 0.14 & 0.2 \\
\jclipII & 0.38 & 0.37 & 0.32\\
\JEmbeddingVFour & 0.71 & 0.72 & 0.56 \\
\bottomrule
\end{tabular}
}
\caption{Comparison of cross-modal alignment scores on 1K of random samples from each dataset.}
\label{table:cross-modal-alignment}
\end{table}

\subsection{Cone Effect} 

\citet{mind_the_gap} demonstrate that multimodal models trained with contrastive loss suffer from an inductive bias known as the {cone effect}. Each modality tends to cluster together in randomized embedding spaces before training, and contrastive loss tends to make the cross-modal matching pairs form a kind of high-dimensional cone, linking one part of the embedding space to another rather than distributing embeddings evenly.

\begin{figure}
  \begin{subfigure}[b]{0.5\textwidth}
  \includegraphics[width=\linewidth]{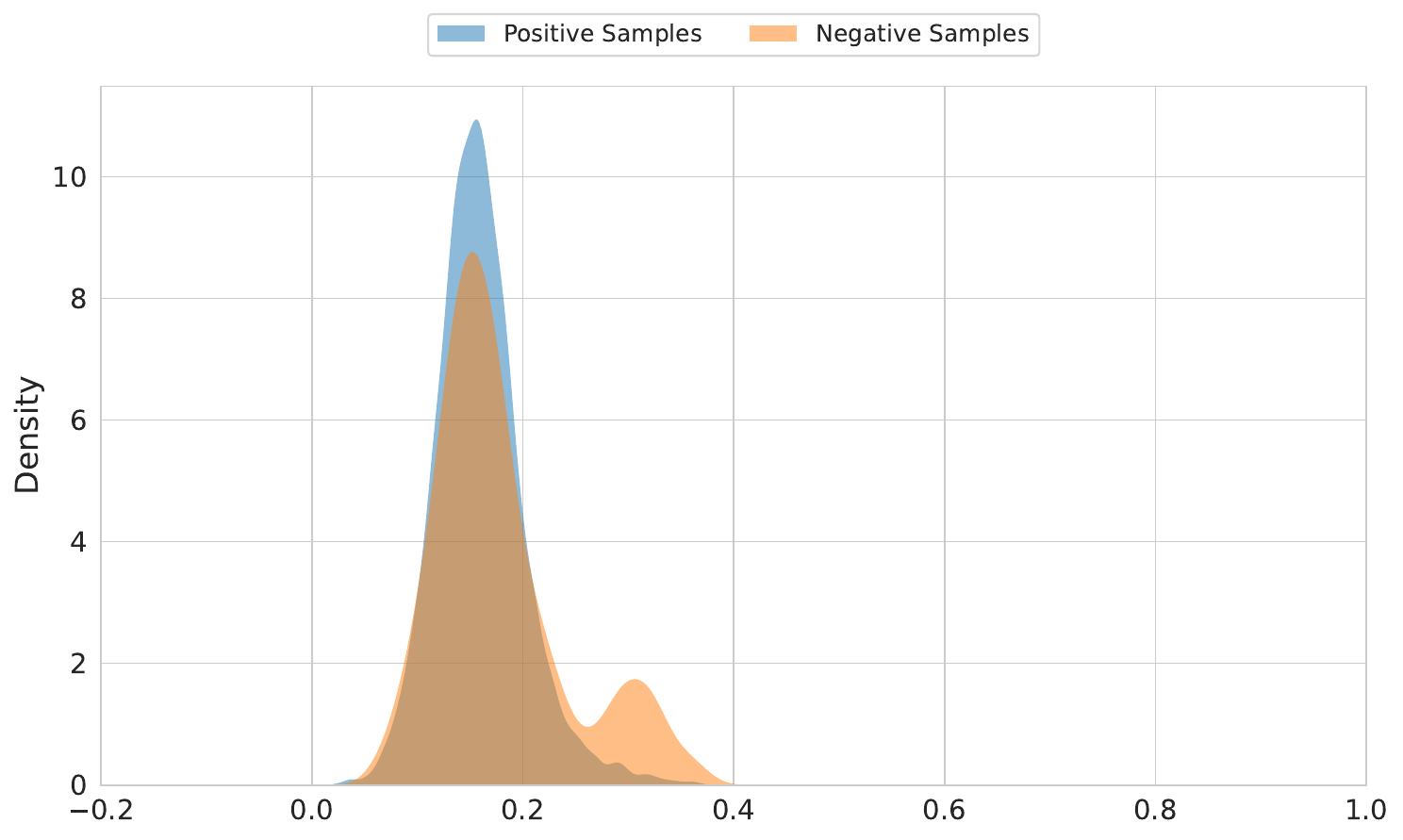}
  \end{subfigure}
  \begin{subfigure}[b]{0.5\textwidth}
  \includegraphics[width=\linewidth]{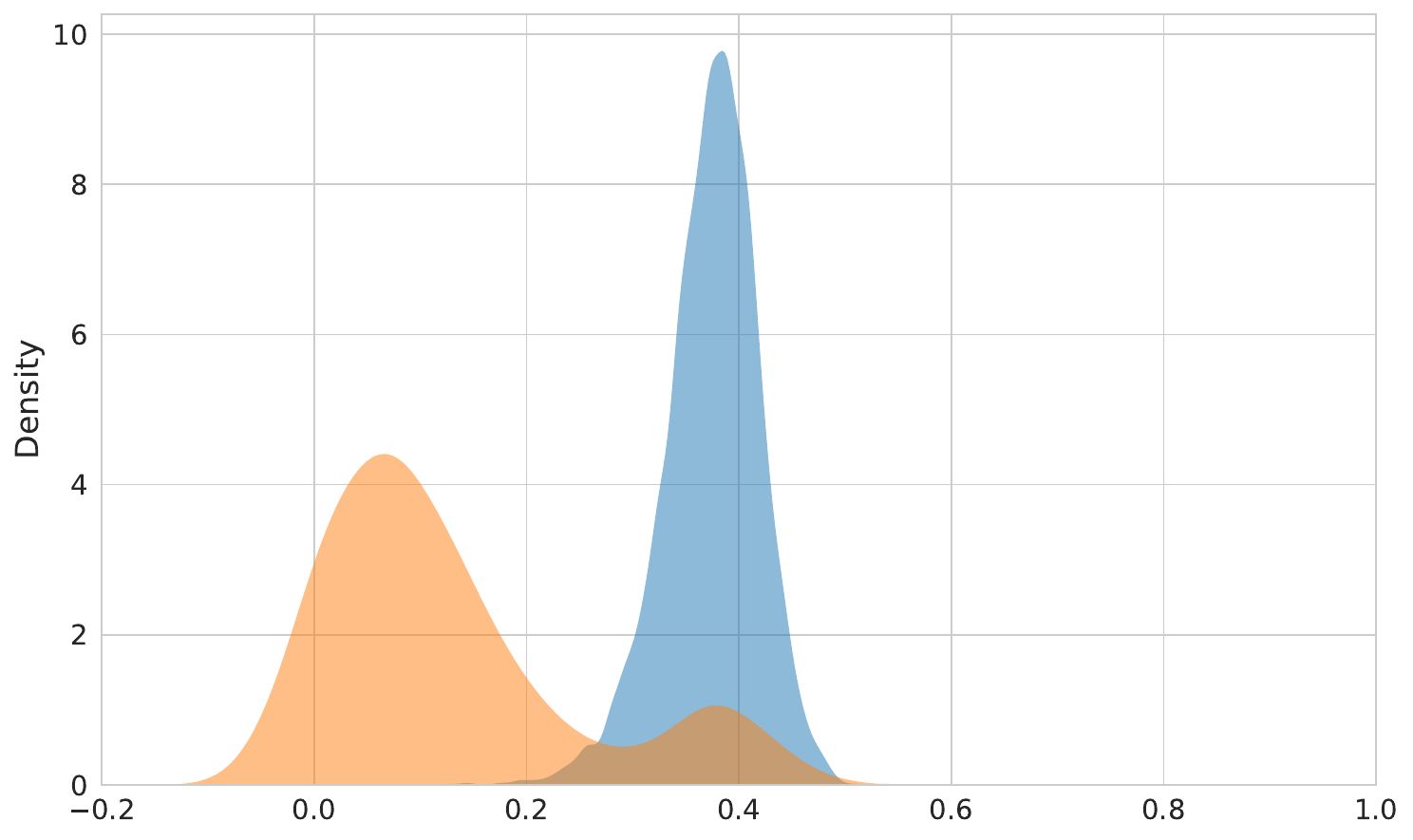}
  \end{subfigure}
  \begin{subfigure}[b]{0.5\textwidth}
  \includegraphics[width=\linewidth]{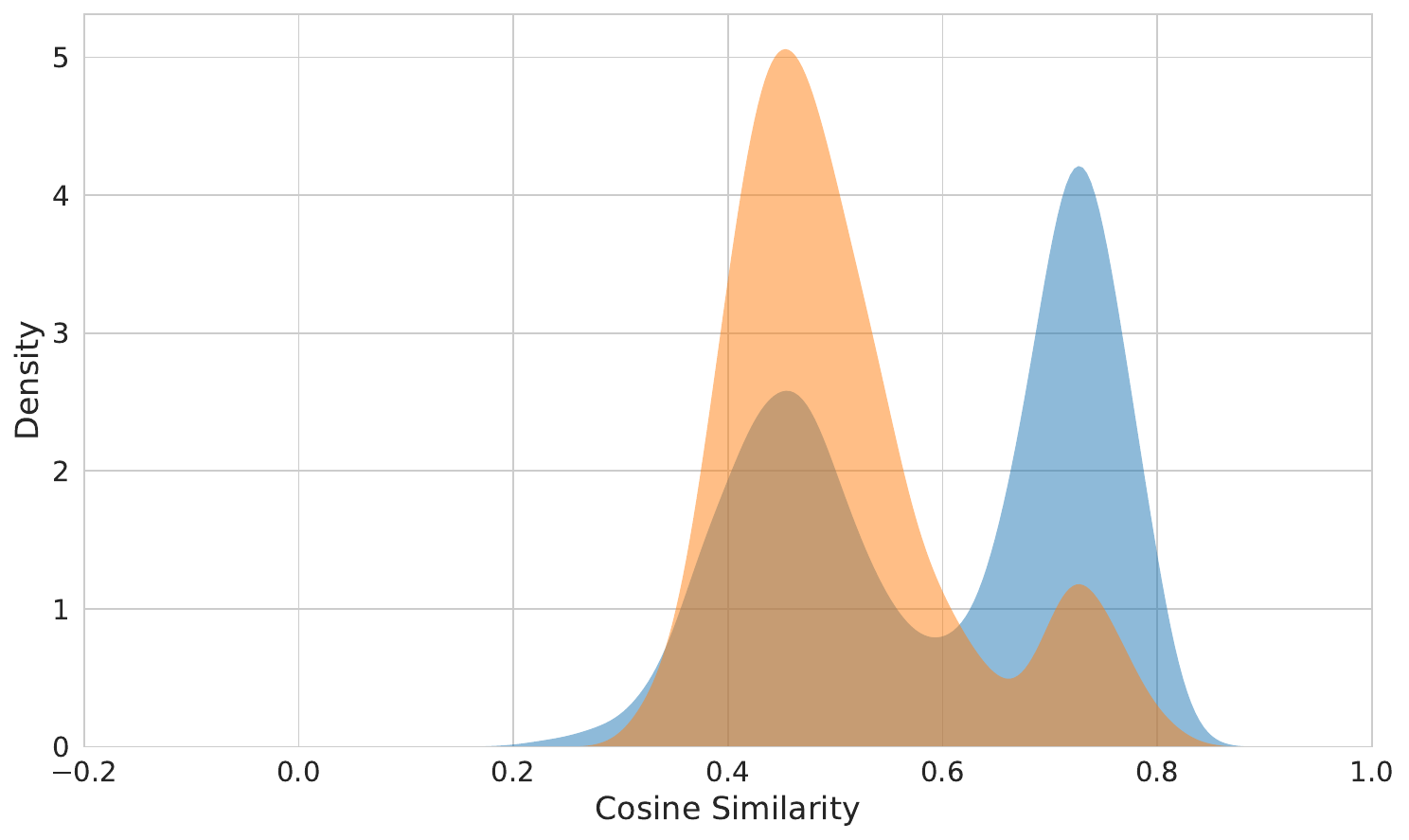}
  \end{subfigure} 
  \caption{Distribution of the cosine similarities of positive (correct matches) versus negative (incorrect matches) image-text samples. (top) OpenAI CLIP, (middle) \jclipII, (bottom) \JEmbeddingVFour.}
  \label{fig:positive-negatives}
\end{figure}

The impact of the cone effect can be seen in Figure~\ref{fig:positive-negatives}. The difference in cosine similarity between correct and incorrect text-image matches is quite small for OpenAI CLIP (top), significantly greater in \jclipII{} (middle), but \JEmbeddingVFour{} (bottom) shows a much greater spread of cosine similarity ranges with very distinctly separate peaks for positive and negative pairs. This shows that \JEmbeddingVFour{} uses much more of the embedding space and image and text embeddings have a much greater overlap in distribution.

\section{Conclusion}
\label{sec:conclusion}

We present \JEmbeddingVFour, a state-of-the-art multimodal and multilingual embedding model designed for a wide range of tasks, including semantic text retrieval, text-to-image retrieval, text-to-visually-rich document retrieval, and code search. The model achieves strong performance using single-vector representations and demonstrates even greater effectiveness with multi-vector representations, particularly in visually rich document retrieval.
\JEmbeddingVFour{} aligns representations across modalities into a single, shared semantic space, sharply reducing structural gaps between modalities compared to CLIP-style dual-tower models, enabling more effective cross-modal retrieval.

In future work, we plan to further enhance this model’s multilingual capabilities and explore techniques to create smaller, more efficient variants.

\FloatBarrier

\balance

\bibliographystyle{unsrtnat}
\bibliography{references}  

\clearpage
\pagenumbering{gobble}
\onecolumn

\appendix
\section{Appendix}
\setcounter{table}{0}
\renewcommand{\thetable}{A\arabic{table}}

\subsection{Datasets in the \JVDR{} Benchmark}
\label{app:dataset-collection}
\small{
\begin{longtable}{L{4.3cm} L{1.8cm} L{1.7cm} L{1.5cm} L{2.0cm} L{1.7cm}}
\caption{Overview of the Dataset Collection} \label{tab:dataset_collection} \\
\toprule
\textbf{Dataset Name} & \textbf{Domain} & \textbf{Document Format} & \textbf{Query Format} & \textbf{Number of Queries / Documents} & \textbf{Languages} \\
\midrule
\endfirsthead

\multicolumn{6}{c}%
{\tablename\ \thetable\ -- \textit{continued from previous page}} \\
\toprule
\textbf{Dataset Name} & \textbf{Domain} & \textbf{Document Format} & \textbf{Query Format} & \textbf{Number of Queries / Documents} & \textbf{Languages} \\
\midrule
\endhead

\midrule \multicolumn{6}{r}{\textit{Continued on next page}} \\
\endfoot

\bottomrule
\endlastfoot

jinaai/airbnb-synthetic-retrieval\dag & Housing & Tables & Instruction & 4953 / 10000 & ar, de, en, es, fr, hi, hu, ja ru, zh \\
jinaai/arabic\_chartqa\_ar & Mixed & Charts & Question & 745 / 342 & ar \\
jinaai/arabic\_infographicsvqa\_ar & Mixed & Illustrations & Question & 120 / 40 & ar \\
jinaai/automobile\_catalogue\_jp & Marketing & Catalog & Question & 45 / 15 & ja \\
jinaai/arxivqa & Science & Mixed & Question & 30 / 499 & en \\
jinaai/beverages\_catalogue\_ru & Marketing & Digital Docs & Question & 100 / 34 & ru \\
jinaai/ChartQA & Mixed & Charts & Question & 996 / 834 & en \\
jinaai/CharXiv-en & Science & Charts & Question & 999 / 1000 & en \\
jinaai/docvqa & Mixed & Scans & Question & 39 / 499 & en \\
jinaai/donut\_vqa & Medical  & Scans / Handwriting & Question & 704 / 800 & en \\
jinaai/docqa\_artificial\_intelligence & Software / IT & Digital Docs & Question & 70 / 962 & en \\
jinaai/docqa\_energy & Energy & Digital Docs & Question & 69 / 971 & en \\
jinaai/docqa\_gov\_report & Government & Digital Docs & Question & 77 / 970 & en \\
jinaai/docqa\_healthcare\_industry & Medial & Digital Docs & Question & 90 / 961 & en \\
jinaai/europeana-de-news & Historic & Scans / News Articles & Question & 379 / 137 & de \\
jinaai/europeana-es-news & Historic & Scans / News Articles & Question & 474 / 179 & es \\
jinaai/europeana-fr-news & Historic & Scans / News Articles & Question & 237 / 145 & fr \\
jinaai/europeana-it-scans & Historic & Scans & Question & 618 / 265 & it \\
jinaai/europeana-nl-legal & Legal & Scans & Question & 199 / 244 & nl \\
jinaai/github-readme-retrieval-multilingual\dag & Software / IT & Markdown Docs & Description & 16953 / 16998 & ar, bn, de, en, es, fr, hi, id, it, ja, ko, nl pt, ru, th, vi, zh \\
jinaai/hindi-gov-vqa & Governmental & Digital Docs & Question & 454 / 337 & hi \\
jinaai/hungarian\_doc\_qa\_hu & Mixed & Digital Docs & Question & 54 / 51 & hu \\
jinaai/infovqa & Mixed & Illustrations & Question & 363 / 500 & en \\
jinaai/jdocqa & News & Digital Docs & Question & 744 / 758 & ja \\
jinaai/jina\_2024\_yearly\_book & Software / IT &Digital Docs & Question & 75 / 33 & en \\
jinaai/medical-prescriptions & Medical & Digital Docs & Question & 100 / 100 & en \\
jinaai/mpmqa-small & Manuals & Digital Docs & Question & 155 / 782 & en \\
jinaai/MMTab & Mixed & Tables & Fact & 987 / 906 & en \\
jinaai/openai-news & Software / IT & Digital Docs & Question & 31 / 30 & en \\
jinaai/owid\_charts\_en & Mixed & Charts & Question & 132 / 937 & en \\
jinaai/plotqa & Mixed & Charts & Question & 610 / 986 & en \\
jinaai/ramen\_benchmark\_jp & Marketing & Catalog & Question & 29 / 10 & ja \\
jinaai/shanghai\_master\_plan & Governmental & Digital Docs & Question / Key Phrase & 57 / 23 & zh, en \\
jinaai/wikimedia-commons-documents-ml\dag & Mixed & Mixed & Description & 15593 / 15217 & ar, bn, de, en, es, fr, hi, hu, id, it, ja, ko, my, nl, pt, ru, th, ur, vi, zh \\
jinaai/shiftproject & Environmental Documents & Digital Docs & Question & 89 / 998 & fr \\
jinaai/stanford\_slide & Education & Slides & Question & 14 / 994 & en \\
jinaai/student-enrollment & Demographics & Charts & Question & 1000 / 489 & en \\
jinaai/tabfquad & Mixed & Tables & Question & 126 / 70 & fr, en \\
jinaai/table-vqa & Science & Tables & Question & 992 / 385 & en \\
jinaai/tatqa & Finance & Digital Docs & Question & 121 / 270 & en \\
jinaai/tqa & Education & Illustrations & Question & 981 / 393 & en \\
jinaai/tweet-stock-synthetic-retrieval\dag & Finance & Charts & Question & 6278 / 10000 & ar, de, en, es, hi, hu, ja, ru, zh \\
jinaai/wikimedia-commons-maps & Mixed & Maps & Description & 443 / 451 & en \\
\end{longtable}
\dag For multilingual datasets, the total number of queries and documents is the sum across all language-specific splits.
}

\subsection{JinaVDR (Visual Document Retrieval) Benchmark Results}
\label{app:jvdr_results}
\small{

\begin{longtable}{L{3.5cm} L{1.0cm} L{1.0cm} L{1.0cm} L{1.0cm} L{1.5cm} L{1.0cm} L{1.0cm} L{1.0cm}}
\caption{Overview of JinaVDR Results for Various Models} \label{tab:jinavdr_overview} \\
\toprule
\textbf{Task} & \textbf{bm25 + OCR} & \textbf{jev3 + OCR} & \textbf{j-clip-v2} & \textbf{colpali-v1.2}  & \textbf{dse-qwen2-2b-mrl-v1} & \textbf{jev4-single} & \textbf{jev4-multi} \\
\midrule
\endfirsthead

\multicolumn{8}{c}%
{\tablename\ \thetable\ -- \textit{continued from previous page}} \\
\toprule
\textbf{Task} & \textbf{bm25 + OCR} & \textbf{jev3 + OCR} & \textbf{j-clip-v2} & \textbf{colpali-v1.2}  & \textbf{dse-qwen2-2b-mrl-v1} & \textbf{jev4-single} & \textbf{jev4-multi} \\
\midrule
\endhead

\midrule \multicolumn{9}{r}{\textit{Continued on next page}} \\
\endfoot

\bottomrule
\endlastfoot

Average                         & 46.88 & 48.97 & 40.96    & 65.39    & 68.89 & 75.47        & 81.52        \\
\midrule
medical-prescriptions           & 38.18       & 38.12       & 15.68          & 66.22          & 38.86       & 80.95              & 97.69              \\
stanford\_slide                 & 81.78       & 95.28       & 91.48          & 100.0            & 100.0         & 100.0                & 97.16              \\
donut\_vqa                      & 19.39       & ~~2.59        & ~~1.46           & 34.12          & 25.31       & 78.60               & 74.08              \\
table-vqa                       & 55.22       & 63.04       & 36.34          & 80.98          & 85.70        & 86.57              & 89.21              \\
ChartQA                         & 28.39       & 31.47       & 39.73          & 54.45          & 58.38       & 70.88              & 71.80               \\
tqa                             & 50.11       & 24.40        & 27.80           & 63.03          & 65.35       & 65.44              & 68.46              \\
openai-news                     & 76.63       & 87.30        & 70.05          & 94.81          & 93.75       & 93.97              & 96.43              \\
europeana-de-news               & 11.26       & 12.02       & 11.18          & 35.20           & 44.32       & 48.89              & 63.76              \\
europeana-es-news               & 51.99       & 43.82       & 12.95          & 45.70           & 60.66       & 60.81              & 80.70               \\
europeana-it-scans              & 39.11       & 38.77       & 16.54          & 58.70           & 54.28       & 58.01              & 73.29              \\
europeana-nl-legal              & 39.38       & 34.24       & 11.30           & 39.13          & 33.12       & 42.77              & 59.82              \\
hindi-gov-vqa                   & ~~1.83        & ~~7.51        & ~~5.21           & 11.43          & 10.19       & 15.32              & 22.49              \\
automobile\_catalogue\_jp       & 20.92       & 50.39       & 32.54          & 35.72          & 66.44       & 72.22              & 81.32              \\
beverages\_catalogue\_ru        & 11.05       & 14.09       & 39.66          & 68.47          & 80.32       & 85.68              & 87.73              \\
ramen\_benchmark\_jp            & 28.02       & 63.37       & 41.28          & 52.03          & 51.66       & 90.77              & 94.65              \\
jdocqa                 & ~~1.64        & ~~7.85        & 19.94          & 35.68          & 67.00          & 75.63              & 82.42              \\
hungarian\_doc\_qa              & 34.28       & 57.84       & 50.44          & 68.83          & 55.25       & 74.64              & 75.56              \\
arabic\_chartqa\_ar             & ~~9.32        & ~~8.63        & ~~6.62           & 26.92          & 49.35       & 62.16              & 66.64              \\
arabic\_infographicsvqa\_ar     & 13.26       & 13.43       & 50.36          & 34.76          & 71.72       & 85.38              & 93.21              \\
owid\_charts\_en                & 66.19       & 62.10        & 57.71          & 78.17          & 84.26       & 92.06              & 92.29              \\
arxivqa                         & 56.73       & 54.41       & 83.41          & 92.54          & 93.33       & 95.44              & 95.44              \\
docvqa                          & 81.11       & 50.81       & 45.29          & 90.38          & 86.28       & 83.06              & 92.98              \\
shiftproject                    & 62.42       & 70.25       & 31.85          & 75.18          & 78.54       & 82.55              & 91.13              \\
docqa\_artificial\_intelligence & 91.68       & 82.98       & 66.52          & 96.09          & 97.52       & 96.43              & 98.04              \\
docqa\_energy                   & 89.97       & 76.97       & 65.56          & 96.03          & 90.08       & 88.66              & 96.28              \\
docqa\_gov\_report              & 87.20        & 82.72       & 68.84          & 92.92          & 94.19       & 92.03              & 95.97              \\
docqa\_healthcare\_industry     & 86.44       & 86.88       & 68.13          & 93.14          & 96.14       & 94.62              & 97.51              \\
tabfquad                        & 45.67       & 80.49       & 47.04          & 89.18          & 92.38       & 95.57              & 95.38              \\
mpmqa\_small                    & 85.54       & 67.39       & 59.72          & 88.88          & 81.62       & 80.44              & 91.28              \\
jina\_2024\_yearly\_book      & 87.67       & 85.98       & 77.12          & 95.77          & 93.39       & 94.29              & 98.17              \\
wikimedia-commons-maps          & ~~5.37        & ~~5.06        & 20.67          & 27.46          & 33.06       & 40.23              & 53.45              \\
plotqa                          & 61.13       & 51.44       & 24.05          & 70.58          & 75.99       & 77.48              & 78.75              \\
MMTab                           & 74.82       & 74.06       & 44.54          & 84.66          & 86.04       & 86.08              & 90.03              \\
CharXiv-en                      & 46.85       & 41.47       & 56.28          & 79.64          & 83.86       & 83.00                 & 87.66              \\
student-enrollment              & ~~1.05        & ~~1.30         & ~~0.70            & ~~3.95           & ~~4.09        & ~~8.04               & 11.55              \\
tatqa                           & 75.62       & 49.88       & 44.23          & 82.57          & 80.97       & 80.14              & 92.76              \\
shanghai\_master\_plan          & 12.69       & 92.67       & 75.28          & 88.87          & 92.56       & 95.53              & 97.41              \\
europeana-fr-news               & 24.55       & 23.69       & 16.43          & 30.33          & 38.23       & 36.66              & 50.16              \\
infovqa                         & 73.61       & 75.09       & 63.38          & 87.53          & 92.64       & 92.16              & 96.69   \\
\end{longtable}
}
\begin{flushleft}
    \textbf{Models:} bm25+OCR:~BM25~with~EasyOCR, jev3+OCR:~\JEmbeddingVThree{}~with~EasyOCR, j-clip-v2:~\jclipII{}, colpali-v1.2:~\href{https://huggingface.co/vidore/colpali-v1.2}{ColPALI-v1.2}, dse-qwen2-
2b-mrl-v1:~\href{https://huggingface.co/MrLight/dse-qwen2-2b-mrl-v1}{DSE-QWen2-2b-MRL-V1}, je4-single:~\JEmbeddingVFour{}~single-vector, jev4-multi:~\JEmbeddingVFour{}~multi-vector
\end{flushleft}

\begin{table}[]
\caption{Retrieval performance on ViDoRe (nDCG@10\%).}
\label{tab:vidore_retrieval}
\setlength{\tabcolsep}{2pt}
\centering
\small
\begin{tabular}{@{}lccccccccccc@{}}
\toprule
\textbf{Model} & \textbf{Avg} & \textbf{AQA} & \textbf{DVQA} & \textbf{InfoVQA} & \textbf{Shift} & \textbf{AI} & \textbf{Energy} & \textbf{Gov} & \textbf{Health} & \textbf{TabFQ} & \textbf{TQA} \\
\midrule
OCR + jina-embeddings-v3                     & 26.02 & 26.31 & 12.62 & 32.79 & 14.18 & 22.84 &  27.47 & 31.16 & 45.78 & 44.54 & 2.53 \\
jina-clip-v2                                 & 53.61 & 68.33 & 27.62 & 60.6 & 34.12 & 66.55 & 64.69 & 67.47 & 68.38 & 46.89 & 31.43 \\
voyage-multimodal-3 & 84.20 & 84.90 & 55.60 & 85.40 & 78.70 &  94.50 & 89.50 & 96.00 & 95.10 & 92.80 & 69.90 \\
colpali-v1.2 & 83.90 & 78.00 & 57.20 & 82.80 & 79.10 & 98.10 & 95.20 & 94.80 & 96.70 & 89.70 & 68.10      \\
dse-qwen2-2b-mrl-v1                  & 85.80   & 85.60   & 57.10  & 88.10   & 82.00         & 97.50                   & 92.90  & 96.00              & 96.40               & 93.10    & 69.40   \\
OCR + bm25                                   & 65.50   & 31.60   & 36.80  & 62.90   & 64.30         & 92.80                   & 85.90  & 83.90              & 87.20               & 46.50    & 62.70   \\
siglip-so400m-patch14-384             & 51.40   & 43.20   & 30.30  & 64.10   & 18.70         & 62.50                   & 65.70  & 66.10              & 79.10               & 58.10    & 26.20   \\
jina-embeddings-v4 (dense) & 84.11 & 83.57 & 50.54 & 87.85 & 84.07 & 97.    16 & 91.66 & 91.48 & 94.92 & 94.48 & 65.35 \\
jina-embeddings-v4 (late) & 90.17 & 88.95 & 59.98 & 93.57 & 92.35 & 99.26 & 96.76 & 96.95 & 98.39 & 95.13 & 80.34 \\
\bottomrule
\end{tabular}

\medskip
\begin{flushleft}
\textbf{Tasks:} Avg:~Mean nDCG@10\% over all tasks, AQA:~ArxivQA, Shift:~Shift~Project, DVQA:~DocVQA, InfoVQA:~InfographicVQA, AI:~Artificial~Intelligence, Gov:~Government~Reports, Health:~Healthcare~Industry, TabFQ:~TabFQuad, TQA:~TAT-DQA
\end{flushleft}
\end{table}

\small{
\begin{longtable}{L{3.0cm} L{0.9cm} L{0.9cm} L{0.9cm} L{0.9cm} L{1.2cm} L{0.9cm} L{0.9cm} L{0.9cm}}
\caption{Wikimedia Commons Retreival Benchmark Results} \label{tab:wikicommons_results} \\
\toprule
\textbf{Language} & \textbf{bm25 + OCR} & \textbf{jev3 + OCR} & \textbf{j-clip-v2} & \textbf{colpali-v1.2}  & \textbf{dse-qwen2-2b-mrl-v1} & \textbf{jev4-single} & \textbf{jev4-multi} \\
\midrule
\endfirsthead

\multicolumn{8}{c}%
{\tablename\ \thetable\ -- \textit{continued from previous page}} \\
\toprule
\textbf{Language} & \textbf{bm25 + OCR} & \textbf{jev3 + OCR} & \textbf{j-clip-v2} & \textbf{colpali-v1.2}  & \textbf{dse-qwen2-2b-mrl-v1} & \textbf{jev4-single} & \textbf{jev4-multi} \\
\midrule
\endhead

\midrule \multicolumn{8}{r}{\textit{Continued on next page}} \\
\endfoot

\bottomrule
\endlastfoot

Average         & 21.99 & 37.43 & 48.63 & 33.60 & 58.67 & 66.04 & 75.63 \\
\midrule
Arabic (ar)     & 19.62  & 38.40  & 45.85  & 28.40    & 63.06  & 71.41   & 81.81    \\
Bengali (bn)    & 22.93  & 44.55 & 49.37  & 26.63   & 52.89  & 66.98   & 76.41    \\
German (de)     & 12.74  & 39.58 & 52.87  & 40.36   & 62.99  & 70.21   & 80.86    \\
English (en)    & 36.45  & 45.24 & 56.58  & 64.98   & 70.23  & 73.55   & 81.66    \\
Spanish (es)    & 12.75  & 46.10  & 54.85  & 41.34   & 66.43  & 71.68   & 80.82    \\
French (fr)     & 15.59  & 36.06 & 35.73  & 43.93   & 41.32  & 53.58   & 59.42    \\
Hindi (hi)      & 16.73  & 36.94 & 48.42  & 18.02   & 50.94  & 62.64   & 71.77    \\
Hungarian (hu)  & 25.38  & 33.88 & 44.42  & 12.67   & 52.35  & 65.86   & 76.00   \\
Indonesian (id) & 28.79  & 39.48 & 50.85  & 40.46   & 62.03  & 66.02   & 73.72    \\
Italian (it)    & 19.63  & 37.98 & 49.77  & 34.76   & 60.05  & 63.96   & 73.68    \\
Japanese (jp)   & 21.41  & 30.43 & 44.03  & 28.83   & 63.71  & 66.50    & 77.13    \\
Korean (ko)     & 34.98  & 35.24 & 47.61  & 29.82   & 68.37  & 71.45   & 81.77    \\
Burmese (my)    & 22.84  & 29.45 & 54.36  & 10.28   & 37.61  & 56.58   & 65.01    \\
Dutch (nl)      & 14.90   & 39.89 & 50.40   & 52.29   & 65.09  & 68.58   & 78.94    \\
Portuguese (pt) & 23.32  & 45.85 & 54.28  & 51.30    & 67.53  & 69.04   & 78.85    \\
Russian (ru)    & 16.82  & 38.95 & 49.34  & 31.88   & 64.44  & 68.86   & 80.70     \\
Thai (th)       & 30.00     & 29.64 & 46.25  & 39.13   & 56.41  & 61.68   & 71.02    \\
Urdu (ur)       & 13.64  & 32.73 & 36.52  & ~~9.45    & 38.76  & 49.76   & 62.17    \\
Vietnamese (vi) & 32.40   & 39.80  & 54.59  & 43.72   & 64.62  & 73.30    & 80.24    \\
Chinese (zh)    & 18.82  & 28.41 & 46.45  & 23.82   & 64.51  & 69.23   & 80.58   

\end{longtable}
}

\small{
\begin{longtable}{L{5.2cm} L{0.9cm} L{0.9cm} L{0.9cm} L{0.9cm} L{1.2cm} L{0.9cm} L{0.9cm} L{0.9cm}}
\caption{GitHub Readme Retrieval Benchmark Results} \label{tab:github_results} \\
\toprule
\textbf{Language} & \textbf{bm25 + OCR} & \textbf{jev3 + OCR} & \textbf{j-clip-v2} & \textbf{colpali-v1.2}  & \textbf{dse-qwen2-2b-mrl-v1} & \textbf{jev4-single} & \textbf{jev4-multi} \\
\midrule
\endfirsthead

\multicolumn{8}{c}%
{\tablename\ \thetable\ -- \textit{continued from previous page}} \\
\toprule
\textbf{Language} & \textbf{bm25 + OCR} & \textbf{jev3 + OCR} & \textbf{j-clip-v2} & \textbf{colpali-v1.2} & \textbf{dse-qwen2-2b-mrl-v1} & \textbf{jev4-single} & \textbf{jev4-multi} \\
\midrule
\endhead

\midrule \multicolumn{9}{r}{\textit{Continued on next page}} \\
\endfoot

\bottomrule
\endlastfoot

Average         & 50.11 & 65.14 & 39.06 & 72.91 & 72.24 & 85.57 & 85.69 \\
\midrule
Arabic (ar)     & 27.49       & 27.98       & 31.02       & 55.19 & 55.95       & 75.02       & 75.26       \\
Bengali (bn)    & ~~1.29        & 28.27       & 26.96       & 49.25 & 47.30        & 65.70        & 66.08       \\
German (de)     & 60.11       & 84.58       & 45.46       & 84.15 & 80.62       & 91.09       & 91.35       \\
English (en)    & 87.43       & 91.67       & 48.69       & 91.10  & 90.69       & 96.94       & 97.34       \\
Spanish (es)    & 78.57       & 83.31       & 43.35       & 84.02 & 78.70        & 89.60        & 90.19       \\
French (fr)     & 77.55       & 83.54       & 42.42       & 83.73 & 79.11       & 90.25       & 90.45       \\
Hindi (hi)      & ~~2.72        & 48.08       & 28.55       & 51.22 & 46.49       & 69.31       & 70.98       \\
Indonesian (id) & 78.05       & 82.46       & 38.59       & 79.67 & 74.57       & 88.42       & 88.62       \\
Italian (it)    & 78.83       & 86.54       & 44.26       & 85.31 & 80.81       & 91.76       & 91.41       \\
Japanese (jp)   & 14.46       & 63.20        & 42.02       & 69.02 & 75.42       & 89.74       & 90.80        \\
Korean (ko)     & 40.01       & 35.23       & 37.87       & 64.16 & 68.83       & 87.04       & 86.89       \\
Dutch (nl)      & 76.52       & 86.36       & 43.25       & 84.10  & 82.85       & 92.83       & 91.37       \\
Portuguese (pt) & 80.33       & 84.46       & 43.88       & 85.00    & 80.09       & 91.43       & 91.47       \\
Russian (ru)    & 39.78       & 50.86       & 37.04       & 78.16 & 78.92       & 89.51       & 88.61       \\
Thai (th)       & ~~1.47        & 36.67       & 37.62       & 65.29 & 65.45       & 77.61       & 76.67       \\
Vietnamese (vi) & 66.70        & 79.67       & 37.14       & 70.05 & 68.20        & 86.90        & 86.94       \\
Chinese (zh)    & 40.52       & 54.53       & 35.89       & 60.05 & 74.05       & 81.44       & 82.26 
\end{longtable}
}

\small{
\begin{longtable}{L{4.5cm} L{0.9cm} L{0.9cm} L{0.9cm} L{0.9cm} L{1.2cm} L{0.9cm} L{0.9cm} L{0.9cm}}
\caption{Tweet Stock Retrieval Benchmark Results} \label{tab:tweetstock_results} \\
\toprule
\textbf{Language} & \textbf{bm25 + OCR} & \textbf{jev3 + OCR} & \textbf{j-clip-v2} & \textbf{colpali-v1.2}  & \textbf{dse-qwen2-2b-mrl-v1} & \textbf{jev4-single} & \textbf{jev4-multi} \\
\midrule
\endfirsthead

\multicolumn{8}{c}%
{\tablename\ \thetable\ -- \textit{continued from previous page}} \\
\toprule
\textbf{Language} & \textbf{bm25 + OCR} & \textbf{jev3 + OCR} & \textbf{j-clip-v2} & \textbf{colpali-v1.2}  & \textbf{dse-qwen2-2b-mrl-v1} & \textbf{jev4-single} & \textbf{jev4-multi} \\
\midrule
\endhead

\midrule \multicolumn{8}{r}{\textit{Continued on next page}} \\
\endfoot

\bottomrule
\endlastfoot

Average        & 22.30 & 42.77 & 55.36 & 76.36 & 62.76 & 78.10 & 85.34 \\
\midrule
Arabic (ar)    & ~~0.38                       & ~~1.67                           & 49.36                          & 77.31                          & 52.73                          & 66.15                          & 77.66                          \\
German (de)    & 48.27                      & 66.86                          & 52.49                          & 73.53                          & 57.35                          & 79.38                          & 85.63                          \\
English (en)   & 51.38                      & 63.66                          & 48.35                          & 77.13                          & 63.47                          & 77.92                          & 85.36                          \\
Spanish (es)   & 54.28                      & 63.44                          & 53.44                          & 79.02                          & 62.57                          & 78.68                          & 84.62                          \\
French (fr)    & 51.69                      & 64.76                          & 54.94                          & 76.91                          & 62.17                          & 78.65                          & 85.27                          \\
Hindi (hi)     & ~~0.08                       & ~~0.08                           & 88.55                          & 93.39                          & 97.00                             & 97.46                          & 96.50                           \\
Hungarian (hu) & 15.55                      & 62.31                          & 52.30                           & 71.06                          & 58.17                          & 80.09                          & 85.01                          \\
Japanese (jp)  & ~~0.40                        & 47.80                           & 54.74                          & 70.00                            & 57.76                          & 77.04                          & 85.67                          \\
Russian (ru)   & ~~0.47                       & ~~3.07                           & 47.08                          & 70.72                          & 57.43                          & 76.33                          & 83.11                          \\
Chinese (zh)   & ~~0.45                       & 54.04                          & 52.30                           & 74.54                          & 58.94                          & 69.33                          & 84.55 
\end{longtable}
}

\small{
\begin{longtable}{L{4.5cm} L{0.9cm} L{0.9cm} L{0.9cm} L{0.9cm} L{1.2cm} L{0.9cm} L{0.9cm} L{0.9cm}}
\caption{AirBnB Retrieval Benchmark Results} \label{tab:airbnb_results} \\
\toprule
\textbf{Language} & \textbf{bm25 + OCR} & \textbf{jev3 + OCR} & \textbf{j-clip-v2} & \textbf{colpali-v1.2}  & \textbf{dse-qwen2-2b-mrl-v1} & \textbf{jev4-single} & \textbf{jev4-multi} \\
\midrule
\endfirsthead

\multicolumn{8}{c}%
{\tablename\ \thetable\ -- \textit{continued from previous page}} \\
\toprule
\textbf{Language} & \textbf{bm25 + OCR} & \textbf{jev3 + OCR} & \textbf{j-clip-v2} & \textbf{colpali-v1.2}  & \textbf{dse-qwen2-2b-mrl-v1} & \textbf{jev4-single} & \textbf{jev4-multi} \\
\midrule
\endhead

\midrule \multicolumn{8}{r}{\textit{Continued on next page}} \\
\endfoot

\bottomrule
\endlastfoot

Average        & ~~7.20 & ~~1.13 & ~~2.13 & 10.42 & 11.10 & ~~8.18 & 37.51 \\
\midrule
Arabic (ar)    & ~~1.10                           & ~~0.40                          & ~~0.47                          & ~~3.06                           & ~~3.64                           & ~~2.20                           & ~~6.20                            \\
German (de)    & ~~4.03                          & ~~0.71                         & ~~5.54                          & 20.17                          & 15.09                          & ~~9.27                          & 41.94                          \\
English (en)   & 48.39                         & ~~1.70                          & ~~4.83                          & 23.26                          & 12.94                          & 13.33                         & 64.17                          \\
Spanish (es)   & ~~6.25                          & ~~0.18                         & ~~2.10                           & 18.06                          & ~~8.61                           & ~~9.11                          & 39.84                          \\
French (fr)    & ~~3.86                          & ~~2.00                            & ~~2.05                          & 10.86                          & 11.87                          & ~~8.70                           & 30.55                          \\
Hindi (hi)     & ~~0.16                          & ~~0.86                         & ~~0.82                          & ~~3.19                           & ~~4.93                           & ~~4.05                          & 17.44                          \\
Hungarian (hu) & ~~5.58                          & ~~0.69                         & ~~3.01                          & ~~7.34                           & 11.10                           & ~~6.69                          & 27.30                           \\
Japanese (jp)  & ~~0.36                          & ~~1.53                         & ~~0.54                          & ~~3.44                           & 14.91                          & ~~7.63                          & 45.65                          \\
Russian (ru)   & ~~1.67                          & ~~1.39                         & ~~0.88                          & 13.16                          & 13.61                          & ~~8.66                          & 40.80                           \\
Chinese (zh)   & ~~0.58                          & ~~1.84                         & ~~1.04                          & ~~1.62                           & 14.28                          & 12.14                         & 61.19    
\end{longtable}
}

\newpage

\subsection{CLIP}
\label{app:clip}
\begin{table}[h]
\caption{Cross-modal (Text-to-image) retrieval performance (Recall@5\%) on the CLIP benchmark.}
\label{tab:clip}
\centering
\small
\begin{tabular}{lccccc}
\toprule
\textbf{Model} & \textbf{Avg} & \textbf{flickr30k} & \textbf{mscoco\_captions} & \textbf{crossmodal3600} & \textbf{xtd10} \\
\midrule
nllb-clip-large-siglip  & 83.19 & \textbf{92.24} & 70.84 & \textbf{82.07} & 87.60 \\
jina-clip-v2            & 81.12 & 89.84 & 68.35 & 81.43 & 84.87 \\
jina-embeddings-v4      & \textbf{84.11} & 91.36 & \textbf{76.18} & 79.42 & \textbf{89.46} \\
\bottomrule
\end{tabular}
\medskip
    \\ \textbf{Avg}: Mean Recall@5\% over all 4 tasks.
\end{table}

\begin{table}[h]
\caption{Text-to-image retrieval performance (Recall@5\%) on \textbf{xtd10} for all supported languages.}
\label{tab:xtd10}
\centering
\small
\begin{tabular}{lccc}
\toprule
\textbf{Language} & \textbf{jina-embeddings-v4} & \textbf{jina-clip-v2} & \textbf{nllb-clip-large-siglip} \\
\midrule
\textbf{average} & \textbf{89.46} & 84.87 & 87.60 \\
\midrule
de  & \textbf{92.10} & 85.70 & 88.30 \\
en  & \textbf{93.10} & 89.40 & 89.40 \\
es  & \textbf{91.50} & 85.90 & 88.20 \\
fr  & \textbf{91.30} & 85.10 & 87.70 \\
it  & \textbf{92.20} & 85.80 & 89.30 \\
ko  & \textbf{86.30} & 82.10 & 85.20 \\
pl  & 89.10 & 86.50 & \textbf{89.40} \\
ru  & \textbf{91.50} & 81.10 & 83.40 \\
tr  & 84.70 & 83.70 & \textbf{88.30} \\
zh  & 82.80 & 83.40 & \textbf{86.80} \\
\bottomrule
\end{tabular}
\end{table}

\begin{table}[h]
\caption{Text-to-image retrieval performance (Recall@5\%) on \textbf{crossmodal3600} for all supported languages.}
\label{tab:cm3600}
\centering
\small
\begin{tabular}{lccc}
\toprule
\textbf{Language} & \textbf{jina-embeddings-v4} & \textbf{jina-clip-v2} & \textbf{nllb-clip-large-siglip} \\
\midrule
\textbf{average} & 79.42 & 81.43 & \textbf{82.07} \\
\midrule
ar & 75.75 & 73.56 & \textbf{78.92} \\
bn & 57.97 & 63.78 & \textbf{75.19} \\
da & 80.47 & 85.39 & \textbf{87.14} \\
de & \textbf{91.75} & 91.25 & 89.56 \\
el & 66.50 & 75.03 & \textbf{77.83} \\
en & \textbf{76.47} & 75.83 & 73.11 \\
es & \textbf{83.64} & \textbf{83.64} & 82.64 \\
fi & 66.67 & 82.83 & \textbf{86.42} \\
fr & 88.69 & \textbf{88.78} & 87.86 \\
hi & 47.81 & 55.25 & \textbf{60.31} \\
id & \textbf{87.41} & 84.22 & 86.31 \\
it & 87.97 & \textbf{88.33} & 85.94 \\
ja & \textbf{91.22} & 87.03 & 86.06 \\
ko & \textbf{82.19} & 78.81 & 78.75 \\
nl & 81.00 & \textbf{82.56} & 81.69 \\
no & 71.94 & 81.08 & \textbf{82.69} \\
pl & 80.86 & \textbf{84.00} & 82.72 \\
pt & 81.42 & 82.42 & \textbf{82.69} \\
ro & 84.33 & 89.36 & \textbf{90.03} \\
ru & \textbf{90.28} & 88.97 & 86.44 \\
sv & 72.58 & 78.06 & \textbf{79.33} \\
th & \textbf{83.36} & 81.61 & 81.14 \\
tr & 73.08 & 81.31 & \textbf{83.47} \\
uk & 86.28 & \textbf{88.56} & 85.44 \\
vi & \textbf{88.81} & 86.64 & 85.56 \\
zh & \textbf{86.67} & 78.97 & 76.56 \\
\bottomrule
\end{tabular}
\end{table}
\newpage

\subsection{MTEB and MMTEB}
\label{app:mteb}

\begin{table}[ht]
\centering
\caption{Evaluation Results for Various Models on MTEB Retrieval Tasks (nDCG@10\%)}
\label{tab:mteb_retrieval}
\small{
\setlength{\tabcolsep}{4pt}
\begin{tabular}{lccccccccccc}
\toprule
\textbf{Model} & \textbf{Arg} & \textbf{CQG} & \textbf{CQU} & \textbf{CFHN} & \textbf{FEV} & \textbf{FiQA} & \textbf{HPQA} & \textbf{SCI} & \textbf{TREC} & \textbf{TOU}  & \textbf{AVG} \\
\midrule
multilingual-e5-large & 54.36 & 58.70 & 39.89 & 26.00 & 83.79 & 43.82 & 70.55 & 17.45 & 71.15 & 49.59 & 51.53 \\
e5-mistral-7b-instruct & 61.65 & 63.52 & 46.75 & 28.50 & 86.99 & 56.81 & 73.21 & 16.32 & 87.03 & 55.44 & 57.62 \\
text-embedding-3-large & 57.99 & 65.40 & 50.02 & 30.10 & 88.53 & 55.00 & 71.66 & 23.07 & 79.56 & 58.42 & 57.98 \\
gemini-embedding-001 & 86.44 & 70.68 & 53.69 & 31.06 & 88.98 & 61.78 & 87.01 & 25.15 & 86.32 & 52.39 & 64.35 \\
jina-embedding-l-en-v1 & 48.3 & 51.68 & 38.66 & 25.93 & 71.16 & 41.02 & 57.26 & 18.54 & 60.34 & 62.34 & 47.52 \\
jina-embeddings-v2-base-en & 44.18 & 56.52 & 38.66 & 23.77 & 73.41 & 41.58 & 63.24 & 19.86 & 65.91 & 63.35 & 49.05 \\
jina-embeddings-v3\dag & 54.33 & 58.02 & 43.52 & 43.14 & 89.90 & 47.35 & 64.70 & 19.92 & 77.74 & 55.28 & 55.39 \\
jina-embeddings-v4\dag & 67.07 & 57.59 & 42.95 & 34.57 & 87.16 & 46.51 & 69.01 & 21.47 & 80.36 & 52.41 & 55.91 \\
\bottomrule
\end{tabular}
}
\begin{flushleft}
\dag using the text-matching adapter
\medskip
\\ \textbf{Tasks}: Arg:~ArguAna, CQG:~CQADupstackGamingRetrieval, CQU:~CQADupstackUnixRetrieval, \\ CFHN:~ClimateFEVERHardNegatives, FEV:~FEVERHardNegatives, FiQA:~FiQA2018, \\ HPQA:~HotpotQAHardNegatives, SCI:~SCIDOCS, TREC:~TRECCOVID, TOU:~Touche2020Retrieval.v3
\end{flushleft}
\end{table}


\begin{table}[h]
\caption{Evaluation Results for Various Models on MMTEB Retrieval Tasks (nDCG@10\%)}
\label{tab:mmteb_retrieval}
\centering
\small
\setlength{\tabcolsep}{2pt}
\begin{tabular}{@{}lccccccccccccccccccc@{}}
\toprule
\textbf{Model} & \textbf{Avg} & \textbf{AI} & \textbf{Arg} & \textbf{Bel} & \textbf{Cov} & \textbf{Hag} & \textbf{PK} & \textbf{LB} & \textbf{MIR} & \textbf{ML} & \textbf{SD} & \textbf{SQA} & \textbf{SO} & \textbf{TC} & \textbf{STC} & \textbf{TR} & \textbf{TW} & \textbf{Wiki} & \textbf{WG} \\
\midrule
jina-embeddings-v3                & 58.6 & 32.8 & 54.3 & 73.4 & 78.6 & 98.7 & 38.0 & 93.4 & 62.6 & 73.4 & 19.8 & 0.7 & 90.8 & 77.7 & 39.2 & 0.6 & 73.0 & 89.1 & 18.6 \\
jina-embeddings-v4                & 66.5 & 50.2 & 67.1 & 74.3 & 80.2 & 98.8 & 69.8 & 94.8 & 61.2 & 74.9 & 21.5 & 30.2 & 91.9 & 80.4 & 59.5 & 1.3 & 84.4 & 88.5 & 67.3 \\
bge-m3                            & 55.4 & 29.0 & 54.0 & 78.2 & 77.5 & 98.8 & 59.0 & 90.3 & 69.6 & 74.8 & 16.3 & 7.5 & 80.6 & 54.9 & 21.9 & 1.0 & 37.8 & 89.9 & 41.7 \\
Cohere-embed-mult.-v3             & 59.2 & 29.7 & 55.1 & 81.1 & 77.1 & 98.8 & 38.2 & 93.8 & 68.0 & 76.1 & 19.3 & 4.7 & 89.4 & 83.4 & 24.2 & 0.9 & 75.8 & 90.9 & 58.4 \\
gemini-embedding-001              & 68.1 & 48.8 & 86.4 & 90.7 & 79.1 & 99.3 & 38.5 & 96.0 & 70.4 & 84.2 & 25.2 & 10.3 & 96.7 & 86.3 & 51.1 & 3.0 & 98.0 & 94.2 & 60.5 \\
text-embedding-3-large            & 61.1 & 42.0 & 58.0 & 68.8 & 68.4 & 99.1 & 69.8 & 95.2 & 56.9 & 73.2 & 23.1 & 7.4 & 92.4 & 79.6 & 31.1 & 2.1 & 81.4 & 89.2 & 29.1 \\
voyage-3                          & 66.0 & 42.5 & 61.0 & 76.5 & 88.5 & 98.6 & 94.8 & 94.5 & 57.7 & 75.7 & 21.4 & 10.7 & 94.3 & 80.5 & 49.2 & 1.2 & 85.7 & 89.7 & 67.7 \\
voyage-multilingual-2             & --   & 45.0 & 61.8 & --   & --   & 98.9 & 97.0 & 95.9 & --   & --   & 22.5 & 10.2 & --   & 80.1 & --   & 1.4 & 87.3 & --   & 39.1 \\
\bottomrule
\end{tabular}

\medskip
\begin{flushleft}
\textbf{Tasks:} Avg:~Mean nDCG@10\% for all tasks, AI:~AILAStatutes, Arg:~ArguAna, Bel:~BelebeleRetrieval, Cov:~CovidRetrieval, Hag:~HagridRetrieval, PK:~LEMBPasskeyRetrieval, LB:~LegalBenchCorporateLobbying, MIR:~MIRACLRetrievalHardNegatives, ML:~MLQARetrieval, SD:~SCIDOCS, SQA:~SpartQA, SO:~StackOverflowQA, TC:~TREC-COVID, STC:~StatcanDialogueDatasetRetrieval, TR:~TempReasonL1, TW:~TwitterHjerneRetrieval, Wiki:~WikipediaRetrievalMultilingual, WG:~WinoGrande    
\end{flushleft}
\end{table}

\begin{table}[h]
\caption{Retrieval performance on MTEB LongEmbed (nDCG@10\%)}
\label{tab:longembed}
\centering
\small
\begin{tabular}{@{}lcccccccc@{}}
\toprule
\textbf{Model} & \textbf{Avg} & \textbf{NaQA} & \textbf{Needle} & \textbf{Passkey} & \textbf{QMSum} & \textbf{SummScreen} & \textbf{Wikim} \\ 
\midrule
jina-embeddings-v3 & 55.66 & 34.30 & 64.00 & 38.00 & 39.34 & 92.33 & 66.02 \\
jina-embeddings-v4 & 67.11 & 57.52 & 51.75 & 65.50 & 46.49 & 96.30 & 85.08  \\
voyage-multilingual-2 & 79.17 & 64.69 & 75.25 & 97.00 & 51.50 & 99.11 & 87.49 \\
voyage-3 & 74.07 & 54.12 & 57.75 & 94.75 & 51.05 & 97.82 & 88.90 \\
voyage-3-lite & 71.41 & 51.67 & 54.00 & 84.75 & 53.01 & 96.71 & 88.34 \\
bge-m3 & 58.73 & 45.76 & 40.25 & 59.00 & 35.54 & 94.09 & 77.73 \\
text-embedding-3-large & 52.42 & 44.09 & 29.25 & 69.75 & 32.49 & 84.80 & 54.16 \\
Cohere-embed-english-v3 & 42.11 & 25.04 & 30.50 & 38.50 & 23.82 & 75.77 & 59.03 \\
multilingual-e5-large-instruct & 41.76 & 26.71 & 29.50 & 37.75 & 26.08 & 72.75 & 57.79 \\
multilingual-e5-large & 40.44 & 24.22 & 28.00 & 38.25 & 24.26 & 71.12 & 56.80 \\
\bottomrule
\end{tabular}

\medskip
\begin{flushleft}
\textbf{Tasks:} Avg:~Mean nDCG@10\% for all tasks, NaQA:~LEMBNarrativeQARetrieval, Needle:~LEMBNeedleRetrieval, Passkey:~LEMBPasskeyRetrieval, QMSum:~LEMBQMSumRetrieval, SummScreen:~LEMBSummScreenFDRetrieval, Wikim:~LEMBWikimQARetrieval
\end{flushleft}
\end{table}

\begin{table}[h]
\caption{STS performance on MTEB v2 (Spearman correlation \%).}
\label{tab:mteb_sts}
\setlength{\tabcolsep}{3pt}
\centering
\small
\begin{tabular}{l*{10}{c}}
\toprule
\textbf{Model} & \textbf{Avg} & \textbf{BIO} & \textbf{SICK-R} & \textbf{STS12} & \textbf{STS13} & \textbf{STS14} & \textbf{STS15} & \textbf{STS17} & \textbf{STS22} & \textbf{STSB} \\
\midrule
jina-embeddings-v3 & 85.82 & 88.69 & 89.62 & 82.44 & 89.49 & 84.95 & 89.32 & 90.01 & 68.45 & 89.43 \\
jina-embeddings-v4 & 85.89 & 89.21 & 89.23 & 83.50 & 88.61 & 84.77 & 89.69 & 88.71 & 70.71 & 88.58 \\
BAAI/bge-m3 & 80.61 & -- & 79.72 & 78.73 & 79.60 & 79.00 & 87.81 & 87.13 & 67.99 & 84.87 \\
Cohere-embed-English-3 & 82.40 & 83.50 & 81.27 & 74.37 & 85.20 & 80.98 & 89.23 & 90.34 & 68.18 & 88.55 \\
Cohere-embed-multilingual-v3 & 83.05 & 85.01 & 82.18 & 77.62 & 85.16 & 80.02 & 88.92 & 90.09 & 69.63 & 88.79 \\
gemini-embedding-001 & 85.29 & 88.97 & 82.75 & 81.55 & 89.89 & 85.41 & 90.44 & 91.61 & 67.97 & 89.08 \\
multilingual-e5-large & 81.39 & 84.57 & 80.23 & 80.02 & 81.55 & 77.72 & 89.31 & 88.12 & 63.66 & 87.29 \\
text-embedding-3-large & 81.44 & 84.68 & 79.00 & 72.84 & 86.10 & 81.15 & 88.49 & 90.22 & 66.89 & 83.56 \\
voyage-3 & 78.59 & 87.92 & 79.63 & 69.52 & 80.56 & 73.33 & 80.39 & 86.81 & 69.60 & 79.53 \\
voyage-large-2 & 82.63 & 89.13 & 79.78 & 72.94 & 83.11 & 77.21 & 85.30 & 88.77 & -- & 84.78 \\
voyage-multilingual-v2 & 76.98 & 87.11 & 78.97 & 67.30 & 80.09 & 71.98 & 78.07 & 86.52 & 67.02 & 75.79 \\
\bottomrule
\end{tabular}

\medskip
\begin{flushleft}
\textbf{Tasks:} Avg:~Mean Spearman Correlation \% for all tasks,  BIO:~BIOSSES, STS22:~STS22v2, STSB:~STSBenchmark
\end{flushleft}
\end{table}

\begin{table}[h]
\caption{STS performance on MMTEB v2 (Spearman correlation \%).}
\label{tab:mmteb_sts}
\setlength{\tabcolsep}{1.5pt}
\centering
\small
\begin{tabular}{@{}lcccccccccccccccc@{}}
\toprule
\textbf{Model} & \textbf{Avg} & \textbf{Faro} & \textbf{FinPara} & \textbf{Indic} & \textbf{JSICK} & \textbf{SICK-R} & \textbf{STS12} & \textbf{STS13} & \textbf{STS14} & \textbf{STS15} & \textbf{STS17} & \textbf{STS22} & \textbf{STSB} & \textbf{STSES} & \textbf{SemRel} \\
\midrule
jina-embeddings-v4 & 72.70 & 72.28 & 14.43 & 35.22 & 80.33 & 89.23 & 83.50 & 88.61 & 84.77 & 89.69 & 88.71 & 70.71 & 88.58 & 75.31 & 56.46 \\
jina-embeddings-v3 & 75.77 & 80.82 & 22.38 & 54.66 & 78.16 & 89.62 & 82.44 & 89.49 & 84.94 & 89.31 & 85.94 & 71.14 & 89.44 & 77.87 & 64.58 \\
bge-m3 & 72.99 & 77.80 & 30.43 & 52.13 & 79.21 & 79.72 & 78.73 & 79.60 & 79.00 & 87.81 & 79.65 & 70.03 & 84.87 & 77.50 & 65.38 \\
Cohere-embed-mult.-v3 & 73.77 & 75.95 & 28.24 & 46.73 & 77.19 & 82.18 & 77.62 & 85.16 & 80.02 & 88.92 & 90.09 & 69.36 & 88.79 & 78.76 & 63.84 \\
gemini-embedding-001 & 78.35 & 86.12 & 28.60 & 62.87 & 84.99 & 82.75 & 81.55 & 89.89 & 85.41 & 90.44 & 88.58 & 71.69 & 89.08 & 81.75 & 73.14 \\
text-embedding-3-large & 70.17 & 74.96 & 23.51 & 12.59 & 81.24 & 79.00 & 72.84 & 86.10 & 81.15 & 88.49 & 90.22 & 69.29 & 83.56 & 74.20 & 65.25 \\
voyage-3 & 68.33 & 72.51 & 22.51 & 41.63 & 71.76 & 79.63 & 69.52 & 80.56 & 73.33 & 80.39 & 76.24 & 71.88 & 79.53 & 72.51 & 64.66 \\
voyage-multilingual-2 & 68.02 & 74.42 & 27.07 & 35.03 & 75.94 & 78.97 & 67.30 & 80.09 & 71.98 & 78.07 & 77.06 & 69.03 & 75.79 & 76.69 & 64.88 \\
\bottomrule
\end{tabular}

\medskip
\begin{flushleft}
\textbf{Tasks:} Avg:~Mean Spearman Correlation \% for all tasks, Faro:~FaroeseSTS, FinPara:~FinParaSTS, Indic:~IndicCrosslingualSTS, STS22:~STS22v2, STSB:~STSBenchmark, SemRel:~SemRel24STS
\end{flushleft}
\end{table}

\begin{table}[h]
\caption{Performance on MTEB Code Information Retrieval (MTEB-CoIR) (nDCG@10\%).}
\label{tab:code_retrieval}
\setlength{\tabcolsep}{2pt}
\centering
\small
\begin{tabular}{l*{11}{c}}
\toprule
\textbf{Model} & \textbf{Avg} & \textbf{AppsR} & \textbf{CCSN} & \textbf{CodeMT} & \textbf{CodeST} & \textbf{CodeSN} & \textbf{CodeTO} & \textbf{CodeTD} & \textbf{CosQA} & \textbf{StackO} & \textbf{SynSQL} \\
\midrule
jina-embeddings-v2-code & 52.24 & 16.37 & 83.97 & 44.40 & 68.66 & 59.62 & 75.68 & 27.25 & 41.92 & 89.26 & 46.99 \\
jina-embeddings-v3 & 55.07 & 29.01 & -- & 59.67 & 78.14 & 53.18 & 77.37 & 30.91 & 35.34 & 90.79 & 41.27 \\
jina-embeddings-v4 & 71.59 & 76.08 & 84.05 & 70.60 & 85.06 & 83.69 & 89.34 & 44.19 & 31.48 & 93.45 & 70.45 \\
Cohere-embed-English-3 & 51.36 & 13.72 & -- & 47.02 & 74.82 & 52.81 & 65.28 & 31.38 & 30.65 & 89.35 & 57.20 \\
Cohere-embed-mult.-v3 & 54.31 & 31.91 & -- & 42.91 & 74.19 & 57.57 & 70.25 & 30.14 & 32.58 & 89.42 & 59.79 \\
gemini-embedding-001 & 73.11 & 93.75 & 81.06 & 56.28 & 85.33 & 84.69 & 89.53 & 31.47 & 50.24 & 96.71 & 69.96 \\
text-embedding-3-large & 62.36 & 28.37 & -- & 68.92 & 80.42 & 73.18 & 84.25 & 34.23 & 31.00 & 92.44 & 68.45 \\
voyage-3 & 67.23 & 73.03 & -- & 66.69 & 83.02 & 77.87 & 89.92 & 33.92 & 28.70 & 94.34 & 57.56 \\
voyage-code-3 & 77.33 & 93.62 & 89.35 & 93.58 & 90.67 & 90.09 & 94.96 & 38.57 & 34.45 & 97.17 & 62.87 \\
\bottomrule
\end{tabular}

\medskip
\begin{flushleft}
\textbf{Tasks:} Avg:~Mean nDCG@10\% for all tasks, AppsR:~AppsRetrieval, COIR:~COIRCodeSearchNetRetrieval, CodeMT:~CodeFeedbackMT, CodeST:~CodeFeedbackST, CodeSN:~CodeSearchNetCCRetrieval, CodeTO:~CodeTransOceanContest, CodeTD:~CodeTransOceanDL, StackO:~StackOverflowQA, SynSQL:~SyntheticText2SQL
\end{flushleft}
\end{table}

\end{document}